\pdfoutput=1

\documentclass[11pt]{article}

\usepackage[preprint]{acl}

\usepackage{times}
\usepackage{latexsym}

\usepackage{tabularx}
\usepackage{xurl}
\usepackage{appendix}
\usepackage{hyperref}

\usepackage{booktabs}
\usepackage{array}
\usepackage{multirow}
\usepackage{graphicx}

\usepackage[T1]{fontenc}

\usepackage[utf8]{inputenc}

\usepackage{microtype}

\usepackage{inconsolata}

\usepackage{graphicx}
\usepackage{xspace}

\title{Long Input Benchmark for Russian Analysis}

\newcommand{\ruLongBench}{LIBRA\xspace}
\newcommand{\librusecHistory}{LibrusecHistory\xspace}
\newcommand{\librusecMHQA}{LibrusecMHQA\xspace}
\newcommand{\ruSciAbstractRetrieval}{ruSciAbstractRetrieval\xspace}

\newcommand{\ruSciPassageCount}{ruSciPassageCount\xspace}
\newcommand{\ruGsm}{ruGSM100\xspace}
\newcommand{\passkey}{Passkey\xspace}
\newcommand{\passkeyWithoutGarbage}{PasskeyWithLibrusec\xspace}
\newcommand{\matreshkaNames}{MatreshkaNames\xspace}
\newcommand{\matreshkaYesNo}{MatreshkaYesNo\xspace}
\newcommand{\LongContextMultiQ}{LongContextMultiQ\xspace}
\newcommand{\wikimqa}{ru2WikiMultihopQA\xspace}
\newcommand{\ruQasper}{ruQasper\xspace}
\newcommand{\ruTrec}{ruTREC\xspace}
\newcommand{\ruQuality}{ruQuALITY\xspace}
\newcommand{\ruTPO}{ruTPO\xspace}
\newcommand{\ruSciFi}{ruSciFi\xspace}
\newcommand{\ruBabilong}{ruBABILong\xspace}
\newcommand{\qaFirst}{ruBABILongQA1\xspace}
\newcommand{\qaSecond}{ruBABILongQA2\xspace}
\newcommand{\qaThird}{ruBABILongQA3\xspace}
\newcommand{\qaFourth}{ruBABILongQA4\xspace}
\newcommand{\qaFifth}{ruBABILongQA5\xspace}

\author{
 \textbf{Igor Churin\textsuperscript{1}},
 \textbf{Murat Apishev\textsuperscript{1,2}},
 \textbf{Maria Tikhonova\textsuperscript{1}},
 \textbf{Denis Shevelev\textsuperscript{1}},
\\
 \textbf{Aydar Bulatov\textsuperscript{3}},
 \textbf{Yuri Kuratov\textsuperscript{4,3}},
 \textbf{Sergej Averkiev\textsuperscript{1}},
 \textbf{Alena Fenogenova\textsuperscript{1}}
\\
\\
 \textsuperscript{1}SaluteDevices,
 \textsuperscript{2}Ecom.tech,
 \textsuperscript{3}MIPT,
 \textsuperscript{4}AIRI
\\
 \small{
   \textbf{Correspondence:} \href{mailto:igor.churin19@gmail.com}{igor.churin19@gmail.com}
 }
}

\begin{document}
\maketitle
\begin{abstract}
Recent advancements in Natural Language Processing (NLP) have fostered the development of Large Language Models (LLMs) that can solve an immense variety of tasks. One of the key aspects of their application is their ability to work with long text documents and to process long sequences of tokens. This has created a demand for proper evaluation of long-context understanding. To address this need for the Russian language, we propose \ruLongBench (Long Input Benchmark for Russian Analysis), which comprises 21 adapted datasets to study the LLM's abilities to understand long texts thoroughly. 
The tests are divided into four complexity groups and allow the evaluation of models across various context lengths ranging from 4k up to 128k tokens. We provide the open-source datasets, codebase, and public leaderboard for \ruLongBench to guide forthcoming research.
\end{abstract}

\section{Introduction}

Large Language Models (LLMs) have demonstrated impressive abilities in many NLP applications. Interacting with people through free-form text instructions, they serve as versatile tools for multiple scenarios, transforming the landscape of AI systems. One direction where LLM usage is developing rapidly includes tasks requiring long text processing, such as summarization and information extraction, where their applications alleviate the handling of long texts for humans. 

However, until recently, most LLMs had difficulties in handling long sequences of tokens and were only able to work with a limited context length of several thousand tokens. In recent years, new methods have enabled the models to increase their context significantly, empowering them to solve a new variety of tasks. This, in turn, and the community's demand for automatic systems solving such tasks at a good level has created a need for a thorough evaluation of LLM long context understanding.

\begin{figure}[t]
\vspace{-30pt}
\includegraphics[width=\linewidth,alt={The illustration of the \ruLongBench benchmark.}]{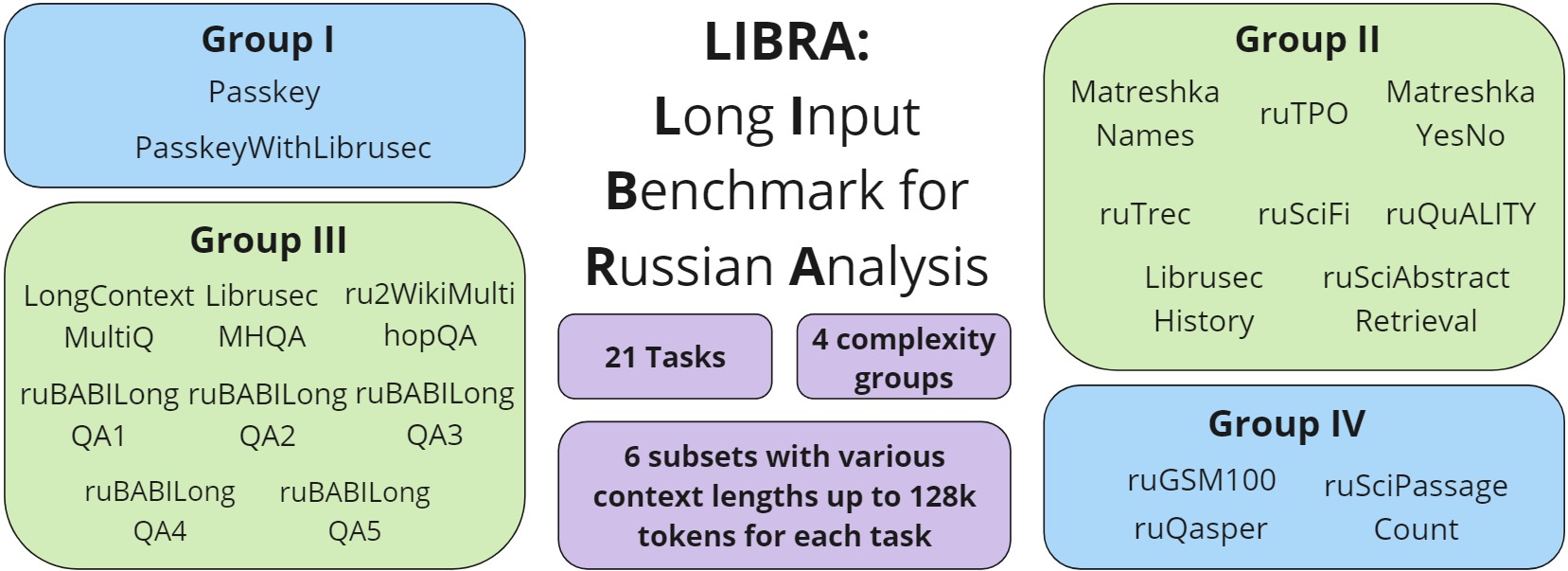}
\caption{\textbf{The LIBRA benchmark} is a set of 21 long-context tasks grouped into four categories based on the complexity of required skills}
\label{fig:libra_idea}
\end{figure}

To address this demand in English, several long context understanding benchmarks have been created recently with LongBench~\cite{bai2023longbench}\footnote{\href{https://huggingface.co/datasets/THUDM/LongBench}{https://huggingface.co/datasets/THUDM/LongBench}} and L-Eval~\cite{an2023eval}\footnote{\href{https://huggingface.co/datasets/L4NLP/LEval}{https://huggingface.co/datasets/L4NLP/LEval}} heading the list. However, the Russian language, at this point, lacks a fair instrument for transparent evaluation of long context understanding.

Our work addresses this problem and presents a new benchmark, which we call \textbf{L}ong \textbf{I}nput \textbf{B}enchmark for \textbf{R}ussian \textbf{A}nalysis, or \ruLongBench, for the evaluation of LLM long context understanding abilities in Russian (see Figure~\ref{fig:libra_idea} for \ruLongBench general structure).

Thus, the contribution of our work can be summarized as follows:
\begin{itemize}
    \item we present a methodology for the evaluation of long-context abilities of LLMs for the Russian language;
    \item we publicly release a set of 21 datasets of various skills and complexities in Russian which form the \ruLongBench benchmark;
    \item we provide a codebase~\footnote{\url{https://github.com/ai-forever/LIBRA}} as long as the number of baseline solutions and public leaderboard\footnote{\url{https://huggingface.co/spaces/ai-forever/LIBRA-Leaderboard}}.
\end{itemize}

\section{Related Work}
\subsection{Long Context Large Language Models}
One of the important tasks in the development of LLMs is to increase the length of the context that the model can understand. This problem has two key points: the complexity of calculations for long sequences and the ability of the model to extract important data in a long context. The solution of the first problem can be attributed to research on the effective processing of the self-attention as in Longformer~\cite{beltagy2020longformer}, LongNet~\cite{ding2023longnet} and FlashAttention~\cite{dao2022flashattention,dao2023flashattention}, using caches for previously calculated outputs such as Transformer-XL~\cite{dai2019transformer}, Unlimiformer~\cite{bertsch2024unlimiformer} and LongLLaMA~\cite{tworkowski2024focused} or replacing it with another mechanism with more effective inference as in RetNet~\cite{sun2023retentive} and Mamba~\cite{gu2023mamba}. The solution to the second problem is to improve positional encoding techniques such as ALiBi~\cite{press2021train} and RoPE-based approaches~\cite{sun2022length, peng2023yarn}.

\subsection{Long Context Benchmarks}
Until recently, most LMs had relatively small context lengths limited by a few thousand tokens. Thus, standard Natural Language Understanding (NLU) benchmarks~\cite{wang2018glue, wang2019superglue, shavrina2020russiansuperglue} contained tasks within this size.

Even today, many ``new generation'' benchmarks created recently, such as  HELM~\cite{liang2022holistic},
\mbox{MT-Bench}~\cite{zheng2023judging}, and Russian-oriented benchmark MERA~\cite{fenogenova2024mera} follow this pattern, limiting their tasks by relatively small context window size to simplify the evaluation procedure and reducing its cost.

The pioneers of long context processing benchmarks have been ZeroSCROLLS~\cite{shaham2023zeroscrolls}\footnote{\href{https://www.zero.scrolls-benchmark.com/}{https://www.zero.scrolls-benchmark.com/}}, designed to test zero-shot model capabilities for NLU over long texts; L-eval~\cite{an2023eval}\footnote{\href{https://huggingface.co/papers/2307.11088}{https://huggingface.co/papers/2307.11088}}, focused on a standardized evaluation methodology for long context LMs addressing two key aspects: dataset construction and evaluation metrics; and LongBench~\cite{bai2023longbench}, the bilingual multi-task benchmark for long context understanding, comprising 21 tasks in English and Chinese. The tasks in LongBench can be divided into 6 big categories and cover key long-text application scenarios, including multi-document QA, single-document QA, summarization, few-shot learning, code completion, and synthesis tasks. 

However, the limitation of the long context benchmarks mentioned above is that they are mainly oriented at the English language (and the Chinese language for LongBench). As for the Russian language, there is an urgent need for a reliable system able to evaluate LLM long context understanding abilities. To address this problem, we propose \ruLongBench, which brings a methodology and 21 tasks for a long context understanding evaluation in Russian.

\section{\ruLongBench}
\label{sec:bench}

\subsection{Benchmark Overview}
\label{sec:bench_overview}

In this section, we introduce {\ruLongBench} (Long Input Benchmark for Russian Analysis), a new benchmark for long context understanding in Russian, which includes 21 tasks for LLM evaluation. \ruLongBench aims to evaluate a large scope of LLMs, including pretrain models and models with supervised finetuning (SFT) with any system prompt that can be picked up.

\begin{table*}
  \centering
  \small
  \renewcommand\arraystretch{0.95}
  \begin{tabular}{@{}cp{0.18\linewidth}lllcr@{}}
    \toprule
    &\textbf{Task Name} & \textbf{Data Origin} & \textbf{Skills} &\textbf{Metric} &\textbf{Dataset Size} \\
    \midrule
    \multirow{2}{*}{\rotatebox[origin=c]{90}{\textbf{I}}}
        & {\passkey}& Translated & Reasoning & EM & 1200\\
        & {\passkeyWithoutGarbage}& New & Reasoning & EM & 1200\\
    \midrule
    \multirow{8}{*}{\rotatebox[origin=c]{90}{\textbf{II}}}
        &{\matreshkaNames} & New & Dialogue Context, Reasoning & EM & 900 \\
        &{\matreshkaYesNo} & New & Dialogue Context, Reasoning & EM & 1799 \\
        &{\librusecHistory}& New & Reasoning & EM & 128\\
        &{\ruTrec}& Translated & Reasoning & EM &  300\\
        &{\ruSciFi} & Translated & World Knowledge, Reasoning & EM & 64\\
        &{\ruSciAbstractRetrieval}& New & Reasoning & EM & 1240\\
        &{\ruTPO} & Translated & Exam, Reasoning & EM & 251 \\
        &{\ruQuality} & Translated & Reasoning & EM & 202\\
    \midrule
    \multirow{8}{*}{\rotatebox[origin=c]{90}{\textbf{III}}}
        &{\LongContextMultiQ}& New & Reasoning & EM & 1200 \\
        &{\librusecMHQA} & New & Reasoning & EM & 384\\
        &{\wikimqa}& Translated & Reasoning & EM & 300 \\
        &{\qaFirst} & Adapted & Reasoning & EM & 600\\
        &{\qaSecond} & Adapted & Reasoning & EM & 600\\
        &{\qaThird} & Adapted & Reasoning & EM& 600 \\
        &{\qaFourth} & Adapted & Reasoning & EM& 600 \\
        &{\qaFifth}& Adapted & Reasoning & EM & 600\\
    \midrule
    \multirow{3}{*}{\rotatebox[origin=c]{90}{\textbf{IV}}} 
        &{\ruSciPassageCount} & New & Reasoning & EM & 600 \\
        &{\ruQasper}& Translated & Reasoning & F1 & 203\\
        &{\ruGsm}& Translated & Math, Logic & EM & 100\\

     
    \bottomrule
  \end{tabular}
  \caption{The {\ruLongBench} tasks outline. The numbers \textbf{I}, \textbf{II}, \textbf{III}, and \textbf{IV} in the left column indicate the complexity group of the tasks described in Subsection~\ref{sec:complexity}. The \textbf{Skills} column defines the skills to be tested on a specific task. \textbf{Data Origin} discloses the source of the dataset.
    The \textbf{Dataset Size} column shows the number of items in the whole dataset.
}
    \label{tab:tasks_info}
\end{table*}

The main purpose of the benchmark is to create a reliable instrument for the long context understanding evaluation, enabling the study of the model's ability to solve various tasks of different complexity with respect to the input context length. For this purpose, all tasks in the \ruLongBench benchmark are divided into 4 complexity groups, and the datasets have several subsets of various context lengths ranging from 4k up to 128k tokens\footnote{See explanation on token length calculation in Section~\ref{sec:context_length}}. The latter makes it possible to explore the influence of the context length on the model results.

\subsection{Complexity group description}
\label{sec:complexity}

In this section, we describe each of the complexity groups of tasks.

\textbf{The first complexity group (I)} consists of tasks that require finding a short text fragment in long textual paragraphs containing irrelevant information. This group includes {\passkey} and {\passkeyWithoutGarbage} datasets. 

\textbf{The second complexity group (II)} includes tasks that require answering the question based on a relevant context. The following types of tasks are related to this group: question answering (QA) such as {\matreshkaNames}, {\matreshkaYesNo}, {\librusecHistory}, {\ruTrec}, {\ruSciFi}, {\ruSciAbstractRetrieval} and multiple choice QA tasks, which are presented by {\ruTPO} and {\ruQuality}. 

The natural development of tasks from the second class of complexity are tasks with questions, the answers to which are not explicitly contained in the text but require the analysis of fragments of input data and the generation of an answer based on it. Such tasks in our classification belong to \textbf{the third complexity group (III)} and represent a multi-hop question answering (MHQA) type. This group includes the following tasks: {\qaFirst}, {\qaSecond}, {\qaThird}, {\qaFourth}, {\qaFifth}, {\LongContextMultiQ}, {\librusecMHQA} and {\wikimqa}. 

Finally, to \textbf{the fourth complexity group (IV)} belongs to the tasks that require understanding the whole context, solving mathematical problems, and QA tasks within complex domains. This group includes {\ruSciPassageCount}, {\ruGsm} and {\ruQasper} datasets.

It should also be mentioned that we do not include code generation and analysis tasks in {\ruLongBench} as most of the software code in the world is written in languages based on English. 

\subsection{Context Length Estimation}
\label{sec:context_length}
In the {\ruLongBench} benchmark, we divide all datasets into subsets of various context lengths. We measure context length in tokens; however, it may vary across different models and tokenizers. In our work, we used the fertility of tokenizers to distribute samples across different context lengths, which indicates the average number of tokens in which one word is tokenized. Thus, the average length in tokens for the text can be approximated by the number of words multiplied by the fertility number.

For the fertility approximation, we calculate the average fertility of the classic LLM tokenizers,  which we further evaluate as baselines (see Subsection~\ref{sec:baselines} for model description) on a complete list of datasets. The fertility of each model is shown in Table ~\ref{tab:fertility}. The average fertility is 2.8. However, we decided to choose it with a margin so that the multilingual model with the highest fertility can be tested on the entire benchmark. As a result, we set the standard fertility to 3.
\begin{table}
  \centering
  \tiny
  \renewcommand\arraystretch{0.95}
  \begin{tabular}{lc}
    \toprule
    \textbf{Model Name} & \textbf{Fertility} \\
    \midrule
        GLM4-9B-Chat & 2.15 \\
        Saiga-LLaMA-3-8B & 2.40 \\
        LLaMA-3-8B & 2.40 \\
        LLaMA-3-8B-Instruct & 2.40 \\
        LLaMA-2-7B-32K & 2.83 \\
        LongAlpaca-7B & 2.83 \\
        LongChat-7B-v1.5-32k & 2.83 \\
        Mistral-7B-v0.1 & 3.08\\
        Mistral-7B-v0.3 & 3.08 \\
        Mistral-7B-Instruct-v0.3 & 3.08\\
        ChatGLM2-6B-32k & 3.50 \\
    \bottomrule
  \end{tabular}
  \caption{The table presents the average model's fertility. \textbf{Model Name} shows the name of a model.
    The \textbf{Fertility} shows the fertility.
}
    \label{tab:fertility}
\end{table}

\begin{table*}
\small
  \centering
  \renewcommand\arraystretch{0.95}
  \begin{tabular}{@{}cp{0.18\linewidth}lcccccc@{}}
    \toprule
    &\textbf{Dataset Name} & \textbf{4k} & \textbf{8k} & \textbf{16k} & \textbf{32k} & \textbf{64k} & \textbf{128k} \\
    & & size / avg len & size / avg len & size/ avg len & size / avg len & size / avg len & size / avg len\\
    \midrule
    \multirow{2}{*}{\rotatebox[origin=c]{90}{\textbf{I}}}
        &{\passkey}& 200 / 2790 & 200 / 5450 & 200 / 10996 & 200 / 21730 & 200 / 43391 & 200 / 87974 \\
        &{\passkeyWithoutGarbage}& 200 / 2705 & 200 / 5563 & 200 / 10835 & 200 / 22215 & 200 / 44682 & 200 / 88189 \\
    \midrule
    \multirow{8}{*}{\rotatebox[origin=c]{90}{\textbf{II}}}
        &{\matreshkaNames} & 150 / 3190 & 150 / 6314 & 150 / 12128 & 150 / 24168 & 150 / 48184 & 150 / 96135 \\
        &{\matreshkaYesNo} & 299 / 3200 & 300 / 6317 & 300 / 12134 & 300 / 24173 & 300 / 48189 & 300 / 96142 \\
        &{\librusecHistory} & - & 32 / 4515 & 32 / 9003 & 32 / 17976 & 32 / 35924 & - \\        
        &{\ruTrec} & 32 / 2870 & 50 / 6292 & 91 / 11886 & 122 / 22357 & - & - \\ 
        &{\ruSciFi} & - & - & - & 36 / 19397 & 28 / 40065 & - \\   
        &{\ruSciAbstractRetrieval} & 210 / 3264 & 210 / 7260 & 210 / 15245 & 210 / 31231 & 200 / 63594 & 200 / 127777 \\     
        &{\ruTPO} & - & 251 / 7651 & - & - & - & - \\ 
        &{\ruQuality} & - & 41 / 6380 & 161 / 12387 & - & - & -\\
    \midrule
    \multirow{8}{*}{\rotatebox[origin=c]{90}{\textbf{III}}}        
       &{\LongContextMultiQ}& 200 / 2940 & 200 / 6360 & 200 / 12240 & 200 / 26572 & 200 / 37482 & 200 / 68239 \\
        &{\librusecMHQA} & - & 384 / 4574 & - & - & - & - \\
        &{\wikimqa} & - & 49 / 6378 & 128 / 11633 & 123 / 25523 & - & - \\
        &{\qaFirst} & 100 / 4002 & 100 / 8001 & 100 / 16002 & 100 / 32001 & 100 / 64002 & 100 / 128001\\
        &{\qaSecond} & 100 / 4002 & 100 / 8001 & 100 / 16002 & 100 / 32001 & 100 / 64002 & 100 / 128001 \\
        &{\qaThird} & 100 / 4011 & 100 / 8010 & 100 / 16011 & 100 / 32010 & 100 / 64011 & 100 / 128010 \\
        &{\qaFourth} & 100 / 4014 & 100 / 8013 & 100 / 16014 & 100 / 32013 & 100 / 64014 & 100 / 128013 \\
        &{\qaFifth} & 100 / 4006 & 100 / 8005 & 100 / 16006 & 100 / 32005 & 100 / 64006 & 100 / 128005 \\ 
    \midrule
    \multirow{3}{*}{\rotatebox[origin=c]{90}{\textbf{IV}}}        
        &{\ruSciPassageCount} & 100 / 3528 & 100 / 7128 & 100 / 13616 & 100 / 27160 & 100 / 53108 & 100 / 105949 \\
        &{\ruQasper} & - & 48 / 5768 & 134 / 11071 & 21 / 25185 & - & - \\
        &{\ruGsm} & - & - & 100 / 9083 & - & - & - \\
    \bottomrule
    \end{tabular}
  \caption{Sizes and average sample lengths for the task subsets of various context lengths. \textbf{Dataset Name} shows the name of the dataset.
    The columns \textbf{4k}, \textbf{8k}, \textbf{16k}, \textbf{32k}, \textbf{64k}, \textbf{128k} show the number of samples and average sample lengths in tokens for the corresponding context length.
}\label{tab:dataset_avg_length}
\end{table*}

Finally, using the selected fertility value, we divided all datasets into subsets of various context lengths ranging from 4k to 128k tokens. The resulting dataset sizes and the average sample context lengths are given in Table ~\ref{tab:dataset_avg_length}.

\subsection{Datasets}
\label{sec:data_collection}

This section describes the datasets and data collection process in detail. We decided to create a combined benchmark that will include 1) translations of English datasets by using Google translator API\footnote{\href{https://pypi.org/project/googletrans/}{https://pypi.org/project/googletrans/}}, 2) adaptations to long input tasks in Russian and 3) entirely new datasets based on open data.

We decided not to generate samples using LLMs and instead used annotators to mark up the samples. This helps reduce bias from using models like GPT-4, which are also part of the assessment. However, it does have some drawbacks, as full annotation can be costly and time-consuming in certain cases.

The exact dataset format can be found in Appendix~\ref{app:examples}.

\vspace{0.5em} \noindent\textbf{\passkey}
The {\passkey} is a synthetic QA dataset based on original passkey dataset from LongLLaMA's GitHub repository\footnote{\href{https://github.com/CStanKonrad/long\_llama/blob/main/examples/passkey.py}{https://github.com/CStanKonrad/long\_llama/blob/main/\\examples/passkey.py}}. The main idea of the task is to extract a relevant piece of code number from a long text fragment that was created by repeating short sentence template containing noise. The model must find this code among the irrelevant information.

\vspace{0.5em} \noindent\textbf{\passkeyWithoutGarbage}
The {\passkeyWithoutGarbage} is a more complicated version of {\passkey} QA dataset, in which we use randomly selected texts from the Librusec dataset as noise to make this dataset more difficult for LLMs.

\vspace{0.5em} \noindent\textbf{\ruGsm}
The {\ruGsm} dataset is a translation of gsm100\footnote{\href{https://huggingface.co/datasets/L4NLP/LEval/viewer/gsm100}{https://huggingface.co/datasets/L4NLP/LEval/\\viewer/gsm100}} one from L-Eval. It contains 100 math problems to be solved using Chain-of-Thought in a few-shot mode. This dataset aims to evaluate the model's reasoning and logical skills in maths. The context for all tasks is a prompt of 16 examples with problem descriptions and answers. 

\vspace{0.5em} \noindent\textbf{\wikimqa}
The {\wikimqa} was created by translating the dataset 2WikiMultihopQA\footnote{\href{https://huggingface.co/datasets/THUDM/LongBench/viewer/2wikimqa\_e}{https://huggingface.co/datasets/THUDM/LongBench/\\viewer/2wikimqa\_e}} from LongBench, which consists of selected samples with a long context from the original multi-hop QA dataset 2WikiMultihopQA~\cite{ho2020constructing}. This Wikipedia-based dataset tests reasoning skills by requiring a model to combine information from multiple texts to answer a question. The format of this dataset, which consists of up to 5-hop questions, makes it difficult for LLMs.

\vspace{0.5em} \noindent\textbf{\ruQasper}
The {\ruQasper} was created by translating the Qasper\footnote{\href{https://huggingface.co/datasets/THUDM/LongBench/viewer/qasper\_e}{https://huggingface.co/datasets/THUDM/LongBench/\\viewer/qasper\_e}} dataset from LongBench, which consists of selected samples with a long context from the original questions answering dataset over academic research papers called Qasper~\cite{dasigi2021dataset}. The goal of the task is to find the answer to the question in one of the parts of the article. The context for samples is drawn from scientific articles to make the task more difficult.

\vspace{0.5em} \noindent\textbf{\ruTrec}
The {\ruTrec} was created by translating the TREC\footnote{\href{https://huggingface.co/datasets/THUDM/LongBench/viewer/trec\_e}{https://huggingface.co/datasets/THUDM/LongBench/\\viewer/trec\_e}} from LongBench. The dataset consists of selected samples with a long context from the original TREC~\cite{li2002learning}. This dataset is a type of few-shot in-context learning, in which the model is given several examples to understand the context, and then it has to answer which topic the question relates to.

\vspace{0.5em} \noindent\textbf{\ruQuality}
The {\ruQuality} was created by translating QuALITY\footnote{\href{https://huggingface.co/datasets/L4NLP/LEval/viewer/quality}{https://huggingface.co/datasets/L4NLP/LEval/\\viewer/quality}} from L-Eval, which consists of selected samples with a long context from the original multiple choice QA dataset called QuALITY~\cite{pang2021quality}. The model must find relevant information in the text and answer by choosing one of the four suggested options.

\vspace{0.5em} \noindent\textbf{\ruTPO}
The {\ruTPO} was created by translating TPO\footnote{\href{https://huggingface.co/datasets/L4NLP/LEval/viewer/tpo}{https://huggingface.co/datasets/L4NLP/LEval/viewer/tpo}} from L-Eval. The original dataset in the L-Eval benchmark consists of 15 samples, that are sourced from the TOEFL Practice Online and the dataset TOEFL-QA~\cite{tseng2016towards}. The TPO is a multiple-choice QA dataset, and, therefore, the model must find relevant information in the text and answer by choosing one of the four suggested options.

\vspace{0.5em} \noindent\textbf{\ruSciFi}
The {\ruSciFi} was created by translating SciFi\footnote{\href{https://huggingface.co/datasets/L4NLP/LEval/viewer/sci\_fi}{https://huggingface.co/datasets/L4NLP/LEval/viewer/\\sci\_fi}} from L-Eval, which consists of selected samples with a long context from the original SFGram\footnote{\href{https://github.com/nschaetti/SFGram-dataset}{https://github.com/nschaetti/SFGram-dataset}} dataset, that contains thousands of science-fiction books, novels and movie information. The dataset aims to test the model's ability to follow contextual knowledge instead of parametric knowledge gained at the pretraining stage. The model needs to answer whether the information provided is true or false based on the information from the context and true or false based on the general world knowledge.

\vspace{0.5em} \noindent\textbf{\matreshkaNames}
To create this dataset, we utilized two sets: Matreshka\footnote{\href{https://huggingface.co/datasets/zjkarina/matreshka}{https://huggingface.co/datasets/zjkarina/matreshka}} and a Russian names\footnote{\href{https://www.kaggle.com/datasets/rai220/russian-cyrillic-names-and-sex/data}{https://www.kaggle.com/datasets/rai220/russian-cyrillic-names-and-sex/data}} dataset. The Matreshka dataset comprises brief interactions involving ``user'' and ``bot'' roles, along with a brief description of the topic being discussed by each participant. To form longer contextual samples, we combined multiple interactions and replaced the names ``user'' and ``bot'' with the pull of names taken from the dataset of Russian names. Subsequently, we randomly selected a topic from the combined interactions and the name of the person discussing that topic. The dataset requires the model to identify the individual who discussed the selected topic. 

\vspace{0.5em} \noindent\textbf{\matreshkaYesNo}
The {\matreshkaYesNo} is based on the two datasets: Matreshka and Russian names, similar to the {\matreshkaNames} dataset. Instead of predicting names in the {\matreshkaNames}, the model is supposed to indicate whether this topic was mentioned in the dialog. The dataset is balanced across answers.

\vspace{0.5em} \noindent\textbf{\LongContextMultiQ}
The {\LongContextMultiQ} is a multi-hop QA long context dataset for Russian that is based on data used for the MultiQ~\cite{taktasheva2022tape}\footnote{\href{https://huggingface.co/datasets/ai-forever/MERA/viewer/multiq}{https://huggingface.co/datasets/ai-forever/MERA/\\viewer/multiq}} dataset creation. The original MultiQ dataset is created by multi-hop dataset generation based on Wikidata\footnote{\href{https://www.wikidata.org/wiki/Wikidata:Introduction}{https://www.wikidata.org/wiki/Wikidata:Introduction}} and Wikipedia, and consists of samples with different length. We selected 200 samples from these generated sources with a long context for each context length.

\vspace{0.5em} \noindent\textbf{\ruBabilong}
We adapted the methodology from~\cite{kuratov2024babilong} to create the Russian Benchmark for Artificial Intelligence for Long ({\ruBabilong})-context evaluation. It contains five long-context reasoning tasks for QA using facts hidden among distractor facts and irrelevant background text. The \textbf{\qaFirst} task requires answering a question about a person's location using a single supporting fact. The \textbf{\qaSecond} and \textbf{\qaThird} tasks introduce the challenge of differentiating subjects and objects, utilizing two and three supporting facts, respectively. The \textbf{\qaFourth} task tackles spatial reasoning through two-argument relations, while the \textbf{\qaFifth} task involves tracking multiple objects to solve the three-argument relation problem. Each task contains 100 samples, scaled to six sequence lengths from 4k to 128k. We obtained the task facts by translating the bAbI dataset~\cite{WestonBCM15}, while the background texts were sampled using books from Librusec.

\vspace{0.5em} \noindent\textbf{\librusecHistory} This dataset was created in question-answering (QA) format using Librusec\footnote{\href{https://huggingface.co/datasets/IlyaGusev/librusec}{https://huggingface.co/datasets/IlyaGusev/librusec}}. 

Each sample in the \librusecHistory dataset includes a text paragraph and a corresponding question. To create tasks with different input lengths, we initially selected large texts from various books in different domains and styles, divided them into fragments of several thousand tokens, and created the annotation (see Appendix~\ref{app:human_details}). These fragments and their respective questions and answers became the dataset's samples. Longer samples, with lengths up to 64,000 tokens, were created by supplementing these fragments with neighboring paragraphs from the original large text on both sides, resulting in longer inputs for the task.

\vspace{0.5em} \noindent\textbf{\librusecMHQA}
This dataset was created in multi-hop Question Answering (QA) format, also using Librusec as a {\librusecHistory}. The main difference between these datasets is that in the {\librusecMHQA} dataset, the necessary information for the answer is distributed in several parts of the context, making the task more difficult and allowing us to evaluate the model's reasoning skills better. The generation procedure for samples of different lengths remains the same.

\vspace{0.5em} \noindent\textbf{\ruSciAbstractRetrieval}
The \ruSciAbstractRetrieval is a QA dataset ideologically similiar to the PassageRetrieval~\cite{bai2023longbench}\footnote{\href{https://huggingface.co/datasets/THUDM/LongBench/viewer/passage\_retrieval\_en}{https://huggingface.co/datasets/THUDM/LongBench/\\viewer/passage\_retrieval\_en}} dataset from LongBench, that aims to evaluate model's reasoning skills. Each element of the dataset consists of a summary description of the topic and a set text paragraphs created from abstracts of scientific articles from ruSciBench\footnote{\href{https://huggingface.co/datasets/mlsa-iai-msu-lab/ru\_sci\_bench }{https://huggingface.co/datasets/mlsa-iai-msu-lab/ru\_sci\_bench }}. The goal is to identify the paragraph where the specified topic is discussed. To create this dataset, we randomly choose some abstracts and generate descriptions of their topics using human annotators to acquire targets. 

\vspace{0.5em} \noindent\textbf{\ruSciPassageCount}
The {\ruSciPassageCount} dataset uses the basic idea of the original PassageCount\footnote{\href{https://huggingface.co/datasets/THUDM/LongBench/viewer/passage\_count}{https://huggingface.co/datasets/THUDM/LongBench/\\viewer/passage\_count}} from LongBench. This QA dataset requires the model to use the full context to solve the problem. To generate the data, we randomly select abstracts from the ruSciBench dataset. We then choose a number of repeats and an ID for the paragraph to repeat. Next, we add the remaining non-repeated paragraphs to the repeated paragraph until we reach the desired context length. The resulting sequence of paragraphs is randomly shuffled. The ground truth for each sample is the number of unique paragraphs.

\section{Evaluation Methodology}

\subsection{Baseline models}
\label{sec:baselines}

We evaluate 12 popular LLMs that feature long context capability, including GPT-4o\footnote{Due to resource constraints, we evaluated GPT-4o on only 10\% of each dataset of our benchmark, including each context length. Therefore, the results may not be precise.}, GLM4-9B-Chat~\cite{zeng2022glm}\footnote{\href{https://huggingface.co/THUDM/glm-4-9b-chat}{https://huggingface.co/THUDM/glm-4-9b-chat}}, ChatGLM2-6B-32k~\cite{zeng2022glm}\footnote{\href{https://huggingface.co/THUDM/chatglm2-6b-32k}{https://huggingface.co/THUDM/chatglm2-6b-32k}}, Saiga-LLaMA-3-8B\footnote{\href{https://huggingface.co/IlyaGusev/saiga\_llama3\_8b}{https://huggingface.co/IlyaGusev/saiga\_llama3\_8b}}, LLaMA-3-8B\footnote{\href{https://huggingface.co/meta-llama/Meta-Llama-3-8B}{https://huggingface.co/meta-llama/Meta-Llama-3-8B}}, LLaMA-3-8B-Instruct\footnote{\href{https://huggingface.co/meta-llama/Meta-Llama-3-8B-Instruct}{https://huggingface.co/meta-llama/Meta-Llama-3-8B-Instruct}}, LLaMA-2-7B-32K\footnote{\href{https://huggingface.co/togethercomputer/LLaMA-2-7B-32K}{https://huggingface.co/togethercomputer/LLaMA-2-7B-32K}}, LongAlpaca-7B\footnote{\href{https://huggingface.co/Yukang/LongAlpaca-7B}{https://huggingface.co/Yukang/LongAlpaca-7B}}, LongChat-7B-v1.5-32k, Mistral-7B-v0.1\footnote{\href{https://huggingface.co/mistralai/Mistral-7B-v0.1}{https://huggingface.co/mistralai/Mistral-7B-v0.1}}, Mistral-7B-v0.3\footnote{\href{https://huggingface.co/mistralai/Mistral-7B-v0.3}{https://huggingface.co/mistralai/Mistral-7B-v0.3}}, Mistral-7B-Instruct-v0.3\footnote{\href{https://huggingface.co/mistralai/Mistral-7B-Instruct-v0.3}{https://huggingface.co/mistralai/Mistral-7B-Instruct-v0.3}}. A detailed information about the baseline models is given in Appendix~\ref{app:model}.  

\subsection{Experimental setup}
Since the tasks themselves are long, in order not to go beyond the context window 
we fixed the evaluation of tasks in zero-shot, except for tasks {\ruTrec} and {\ruGsm} in which the few-shot examples provided as a part of long context input. When the input length of the sample surpasses the maximum model context length, we truncate the input sequence from the right. The baselines were evaluated with greedy decoding (temperature = 1.0, num\_beams = 1, do\_sample = False) for reproducibility. 

For each task, we fixed a natural language prompt unified for all the models (see Appendix~\ref{app:examples} for the exact formulation). The prompts were estimated from an empirical analysis of the tasks through a series of experiments. However, it should be noted that further study of this subject is still required.

We run all the experiments on a double NVIDIA A100 GPU.

\begin{table*}
  \centering
  \small
  \renewcommand\arraystretch{0.95}
  \begin{tabular}{lccccccc}
    \toprule
    \textbf{Model Name} & \textbf{4k} & \textbf{8k} & \textbf{16k} & \textbf{32k} & \textbf{64k} & \textbf{128k} & \textbf{Overall} \\
    \midrule
        GPT-4o & \textbf{73.3} & \textbf{73.1} & \textbf{73.5} & \textbf{62.0} & \textbf{65.3} & \textbf{54.8} & \textbf{70.2}\\
        GLM4-9B-Chat & \underline{61.5} & \underline{59.8} & \underline{53.4} & \underline{50.6} & \underline{48.7} & \underline{43.8} & \underline{52.3} \\
        Mistral-7B-Instruct-v0.3 & 48.3 & 44.7 & 37.3 & 32.3 & - & - & 29.9 \\
        Mistral-7B-v0.3 & 46.6 & 42.9 & 37.9 & 32.8 & - & - & 27.4 \\
        LLaMA-2-7B-32K & 45.2 & 43.7 & 36.6 & 33.0 & - & - & 27.1 \\
        LongChat-7B-v1.5-32k & 38.7 & 36.0 & 30.4 & 24.5 & - & - & 22.1 \\
        ChatGLM2-6B-32k & 28.6 & 24.9 & 22.5 & 14.5 & - & - & 15.7 \\
        LongAlpaca & 26.0 & 22.3 & 18.8 & 13.8 & - & - & 13.7 \\
        LLaMA-3-8B-Instruct & 58.1 & 56.9 & - & - & - & - & 21.9 \\
        Saiga-LLaMA-3-8B & 58.7 & 55.0 & - & - & - & - & 21.0\\
        LLaMA-3-8B & 54.6 & 49.4 & - & - & - & - & 18.4 \\
        Mistral-7B-v0.1 & 47.2 & 42.8 & - & - & - & - & 17.3\\
    \bottomrule
    \end{tabular}
  \caption{The table presents the model evaluation scores for different context lengths. \textbf{Model Name} shows the name of the model.
    The columns \textbf{4k}, \textbf{8k}, \textbf{16k}, \textbf{32k}, \textbf{64k}, \textbf{128k} present evaluation scores averaged over all tasks. The \textbf{Overall} score is obtained by averaging the results over all lengths. The best score is put in bold, the second best is underlined.
} \label{tab:results_by_length}
\end{table*}

\begin{table*}
  \centering
  \tiny
  \renewcommand\arraystretch{0.95}
  \begin{tabular}{lcccccccc}
    \toprule
    \textbf{Model Name} &  \textbf{\passkey} & \textbf{\matreshkaYesNo} & \textbf{\matreshkaNames} & \textbf{\passkeyWithoutGarbage} & \textbf{\librusecHistory} & \textbf{\ruGsm} & \textbf{\ruSciPassageCount} & \textbf{\wikimqa} \\
    \midrule
        GPT-4o & \textbf{100.0} & \textbf{80.0} & \textbf{51.7} & \textbf{100.0} & \textbf{97.5} & \textbf{100.0} & \textbf{35.0} & \textbf{76.7} \\
        GLM4-9B-Chat & \underline{100.0} & \underline{68.0} & \underline{47.3} & \underline{100.0} & \underline{82.0} & 8.0 & 7.5 & \underline{48.8}\\
        Mistral-7B-Instruct-v0.3 &  66.7 & 35.3 & 16.3 & 66.6 & 50.8 & 11.0 & \underline{8.2} & 43.2 \\
        Mistral-7B-v0.3 & 66.7 & 32.0 & 10.0 & 66.7 & 68.0 & 9.0 & 0.0 & 41.0 \\
        LLaMA-2-7B-32K & 66.7 & 33.4 & 3.4 & 65.5 & 40.6 & 7.0 & 4.7 & 37.2 \\
        LongChat-7B-v1.5-32k & 66.5 & 33.4 & 5.9 & 66.0 & 26.6 & 5.0 & 4.8 & 35.2\\
        ChatGLM2-6B-32k & 63.7 & 33.4 & 1.3 & 65.0 & 8.6 & 5.0 & 3.7 & 17.5 \\
        LongAlpaca & 42.4 & 30.5 & 0.4 & 40.6 & 13.3 & 2.0 & 3.8 & 30.3 \\
        LLaMA-3-8B-Instruct & 33.3 & 27.3 & 16.6 & 33.3 & 22.7 & 0.0 & 6.5 & 17.7  \\
        Saiga-LLaMA-3-8B & 33.3 & 28.0 & 15.6 & 33.2 & 24.2 & 0.0 & 3.8 & 17.7  \\
        LLaMA-3-8B & 33.3 & 20.2 & 10.0 & 33.3 & 22.7 & 0.0 & 3.3 & 18.4 \\
        Mistral-7B-v0.1 & 35.0 & 16.8 & 8.1 & 38.3 & 23.4 & \underline{13.0} & 1.3 & 23.0\\
    \bottomrule
    \end{tabular}
  \caption{The table presents the evaluation results. \textbf{Model Name} shows the name of the model.
  The score for each task is averaged by the context length. The best score is put in bold, the second best is underlined.
} \label{tab:results_by_tasks_part1}
\end{table*}

\begin{table*}
  \centering
  \tiny
  \renewcommand\arraystretch{0.95}
  \begin{tabular}{lcccccccc}
    \toprule
    \textbf{Model Name} & \textbf{\LongContextMultiQ} & \textbf{\ruSciAbstractRetrieval} & \textbf{\ruTrec} & \textbf{\ruSciFi} & \textbf{\librusecMHQA} & \textbf{\qaFirst} & \textbf{\qaSecond} & \textbf{\qaThird} \\
    \midrule
        GPT-4o &  \textbf{36.7} & \underline{76.9} & \textbf{75.0} & \textbf{75.0} & \textbf{50.0} & \textbf{78.3} & \textbf{36.7} & \underline{21.4} \\
        GLM4-9B-Chat & 7.8 & \textbf{77.8} & \underline{69.9} & \underline{40.9} & 44.5 & \underline{54.1} & \underline{29.8} & \textbf{22.3} \\
        Mistral-7B-Instruct-v0.3 & 4.8 & 43.6 & 42.5 & 15.3 & 33.6 & 14.3 & 2.8 & 6.0 \\
        Mistral-7B-v0.3 &  5.2 & 30.5 & 5.4 & 0.0 & 39.1 & 37.3 & 16.7 & 15.7 \\
        LLaMA-2-7B-32K & \underline{7.9} & 39.1 & 23.8 & 5.6 & 27.6 & 40.3 & 16.6 & 16.3 \\
        LongChat-7B-v1.5-32k & 3.2 & 41.1 & 7.4 & 2.8 & 24.7 & 17.5 & 7.2 & 4.0 \\
        ChatGLM2-6B-32k & 1.2 & 13.6 & 4.5 & 0.0 & 6.8 & 12.2 & 1.5 & 2.5\\
        LongAlpaca & 0.8 & 23.5 & 0.5 & 1.4 & 7.8 & 3.8 & 0.3 & 3.5  \\
        LLaMA-3-8B-Instruct & 4.9 & 31.4 & 27.4 & 0.0 & \underline{46.1} & 23.7 & 4.1 & 4.5\\
        Saiga-LLaMA-3-8B & 4.8 & 31.7 & 26.3 & 0.0 & 45.1 & 25.4 & 4.4 & 6.1 \\
        LLaMA-3-8B & 7.0 & 30.9 & 19.0 & 0.0 & 41.4 & 20.8 & 7.7 & 9.1 \\
        Mistral-7B-v0.1 & 4.4 & 28.5 & 4.0 & 1.4 & 34.1 & 21.0 & 7.7 & 9.0 \\
    \bottomrule
    \end{tabular}
  \caption{The table presents the evaluation results. \textbf{Model Name} shows the name of the model.
    The score for each task is averaged by the context length. The best score is bold, the second best is underlined.
} \label{tab:results_by_tasks_part2}
\end{table*}

\begin{table*}
  \centering
  \tiny
  \renewcommand\arraystretch{0.95}
  \begin{tabular}{lcccccc}
    \toprule
    \textbf{Model Name} &  \textbf{\qaFourth} & \textbf{\qaFifth} & \textbf{\ruQuality} & \textbf{\ruTPO} & \textbf{\ruQasper} & \textbf{Overall} \\
    \midrule
        GPT-4o & \textbf{79.0} & \textbf{90.0} & \textbf{83.3} & \textbf{100.0} & \textbf{31.7} & \textbf{70.2} \\
        GLM4-9B-Chat & \underline{52.8} & \underline{70.3} & \underline{74.1} & \underline{86.9} & 5.0 & \underline{52.3} \\
        Mistral-7B-Instruct-v0.3 &  27.6 & 37.6 & 30.6 & 66.4 & 5.4 & 29.9 \\
        Mistral-7B-v0.3 & 23.6 & 47.1 & 15.2 & 39.7 & \underline{5.8} & 27.4 \\
        LLaMA-2-7B-32K & 16.7 & 43.0 & 15.5 & 54.3 & 4.7 & 27.1 \\
        LongChat-7B-v1.5-32k & 12.7 & 33.3 & 23.1 & 39.6 & 5.0 & 22.1 \\
        ChatGLM2-6B-32k & 0.6 & 8.8 & 49.2 & 29.0 & 2.6 & 15.7 \\
        LongAlpaca & 0.2 & 29.4 & 44.0 & 6.8 & 2.0 & 13.7 \\
        LLaMA-3-8B-Instruct & 19.6 & 25.3 & 34.6 & 78.1 & 2.2 & 21.9 \\
        Saiga-LLaMA-3-8B & 20.3 & 25.2 & 17.9 & 75.7 & 2.5 & 21.0 \\
        LLaMA-3-8B & 19.1 & 22.6 & 8.5 & 58.2 & 2.2 & 18.5 \\
        Mistral-7B-v0.1 & 12.4 & 23.2 & 17.3 & 39.6 & 2.5 & 17.3 \\
    \bottomrule
    \end{tabular}
  \caption{The table presents the evaluation results. \textbf{Model Name} shows the name of the model.
    The score for each task is averaged by the context length. The \textbf{Overall} score is obtained by averaging the results over each task. The best score is put in bold, the second best is underlined. 
} \label{tab:results_by_tasks_part3}
\end{table*}
\section{Results}

The baseline results with respect to context length are shown in Table ~\ref{tab:results_by_length} and with respect to tasks are shown in Tables~\ref{tab:results_by_tasks_part1}, \ref{tab:results_by_tasks_part2}, \ref{tab:results_by_tasks_part3}. Detailed results for each model are given in Appendix~\ref{app:results}.
Based on the obtained results we can draw the following conclusions for each group of tasks.

\textbf{Group I} The tasks from this group are relatively simple, and almost all models pass them well within their maximum input length. The only exception is the LongAlpaca-7B model.

\textbf{Group II} {\matreshkaYesNo}, turns out to be the most straightforward task in the group, which all models cope with naturally. The {\ruTPO} and {\ruQuality} tasks are of medium complexity; several models achieved good scores in them.

The classic QA task {\librusecHistory} is effectively handled by modern models; however, the quality decreases with the input length increase (e.g. for {\ruSciAbstractRetrieval}). 
Nevertheless, in some cases, a larger context is advantageous, as seen in {\ruTrec}, where increasing the input length helps the model handle the task better because this task is designed in a few-shot format.

The most complex tasks in this group can be considered {\matreshkaNames} and {\ruSciFi}. For the first, several models (e.g., ChatGLM2-6B-32k, LLaMA-2-7B-32K, and LongAlpaca-7B) show low results for any input length. {\ruSciFi} with a 64K context is beyond the capabilities of most models. At the same time, the strongest models (GPT-4o and GLM4-9B-Chat) not only show promising results but also improve the score with the length increase.

\textbf{Group III} For tasks from {\ruBabilong}, an increase in context leads to worse results. {\qaSecond} and {\qaThird} turn out to be significantly more complex than others, which coincides with results from ~\cite{kuratov2024babilong}. 
The length of the context plays a significant role; with its growth, the quality immediately begins to decline for all but the strongest models.

{\librusecMHQA} turns out to be a complex dataset; the maximum quality of the models for solving this problem is only 50 for 8k tokens.

\textbf{Group IV} {\ruSciPassageCount} is the most difficult task created from scratch. All models except GPT-4o handle it poorly, even with a 4K input length; the result's sensitivity to the context's size is high. Besides, all open models fail to cope with {\ruQasper} for complex tasks and domains. A similar result is obtained when measuring the quality of solutions to mathematical problems from {\ruGsm}. Our conclusions are similar to those obtained in~\cite{an2023eval}; the only exception is the LLaMA-2 family of models, which performs worse in our experiments, most likely due to translating tasks into the less familiar Russian language.

\textbf{Overall}, SFT models perform better than the pretrain once. In most cases, an increase in the input length negatively affects the capabilities of all models. The results indicate that our prior division of tasks into groups is highly correlated with their complexity.

\section{Conclusion}
The rapid development of LLMs has posed new challenges for evaluating their ability to process long texts. To address this problem, we have introduced \ruLongBench. This benchmark evaluates LLM long context understanding abilities through 21 long-context textual tasks. 

The tasks enable model evaluation across various context lengths ranging from 4k to 128k tokens based on the analysis of dataset context lengths of the models' tokenizers. Our contribution encompasses a benchmark methodology with open-sourced datasets of different lengths and domains, a codebase for model evaluation, and baseline solution scoring. The datasets are published under the MIT license, and the leaderboard\footnote{\url{https://huggingface.co/spaces/ai-forever/LIBRA-Leaderboard}} is publicly accessible on HuggingFace.

\section*{Limitations}

Although the \ruLongBench was created to solve the absence of the long context benchmark for Russian and provides significant advancements in evaluating language models with long contexts, it still has a number of limitations that need to be acknowledged.

\vspace{0.5em} \noindent\textbf{Data Representation}.
The texts included in the benchmark are gathered from specific domains, which might not cover the full range of Russian language usage. This can raise concerns about data privacy, representation, and potential biases within the benchmark. It is important to consider that dialects, regional variations, and sociolects may not be adequately represented, potentially leading to biased performance metrics. As a result, models may excel in benchmark tasks but struggle with texts outside these domains, limiting their generalization ability. The corpus used for the benchmark may become outdated over time. New words, phrases, and usage patterns could emerge, making the benchmark less relevant for future model evaluations.

\vspace{0.5em} \noindent\textbf{Methodology limitations}.
When creating the datasets, we hypothesized that synthetically augmentation of the context length of the datasets, such as \librusecHistory, would not affect the results. Our experiments show that these tasks are pretty challenging for many models. We made this methodological assumption due to the limitations of human data annotation; it is difficult for people to read large texts and concentrate enough to create questions and search for information within them. This data creation method may result in task errors, particularly when a newly extended text fragment contains conflicting information that could impact the answer. However, we found this approach acceptable due to the increased speed and cost-effectiveness.

The current methodology also restricts the number of tasks, and many of them are translated only due to the high cost of data creation.

\vspace{0.5em} \noindent\textbf{Length context}.
The benchmark focuses on evaluating long contexts, but the definition of ``long context''  can differ based on the application and the model. The chosen context lengths may not be ideal for all usage scenarios, and models could exhibit varying performance. In this paper, we have measured the average fertility of baseline model tokenizers on a full list of datasets from our benchmark to sample different contexts and analyzed the models' results on our datasets across various context lengths. 
LMs with more parameters may inherently perform better, but this does not necessarily reflect improvements in long context understanding.

\vspace{0.5em} \noindent\textbf{Data leakage} is a critical concern for modern benchmarks because current models are trained on a significant amount of text from the Internet. Long context benchmarks are particularly risky, as their texts are based on web sources and books. This could potentially lead to data leakage and inaccurate evaluation. However, creating original long texts from scratch not found on the web is exceptionally costly. As a result, we use open sources to develop our benchmark, acknowledging the potential risks. Nevertheless, we firmly believe this will make a valuable contribution to the Russian community, as no long context datasets are currently available.

\vspace{0.5em} \noindent\textbf{Ethical Considerations}.
The data used in the benchmark was created from open data sources. When annotating the data, we obtained transparent permission from all users and made efforts to maintain the confidentiality and anonymity of participants. As the benchmark develops, ongoing efforts are required to identify and minimize biases in the benchmark datasets and evaluation metrics. The benchmark does not currently contain the datasets covering the ethical or AI safety skill evaluation, but this is a space for future work.

\bibliography{custom}

\appendix
\begin{appendices} 
\section*{\appendixname}


\section{Data Annotation Details}
\label{app:human_details}
The datasets {\librusecHistory}, {\librusecMHQA}, and {\ruSciAbstractRetrieval} were created via the crowd-sourced platform.

In the {\librusecHistory}, annotators were instructed to read a lengthy text and generate four questions based on the text and answer them. Guidelines were provided regarding the type of questions to ask: 1) Questions should be answerable using information present in the text 2) The questions must not be about widely known information but should be related to the text 3) Questions can cover various aspects such as character actions, appearance, thoughts, events, and scene descriptions 4) Logical deductions are not required to answer the questions 5) Each question should have a single, clear, unambiguous answer from the text.

The design of the dataset {\librusecMHQA} project follows a similar structure to {\librusecHistory}, but the question criteria were more complex. In this dataset, the questions were answered by expert editors rather than through crowd-sourcing. The main distinction in the criteria for annotators is the multi-hop questions, where simply reading the sentence containing the answer is insufficient. Instead, reading at least a paragraph of 2-5 sentences, or the entire relevant fragment, is necessary to gather information and generate a complete answer.

The {\ruSciAbstractRetrieval} was collected by crowd-sourced annotators. These annotators were asked to read a long text annotation and briefly describe the contents. The criteria for the description were as follows: 1) The description must start with the word ``Describes''. 2) It must be a single sentence, which can be complex. 3) The description should not exceed 30 words, including conjunctions, particles, and prepositions. 4) It should include the main general ideas identified in the abstract but should not include details.

Training examples were available for all projects.
The contributions of human annotators are amassed and stored in a manner that ensures anonymity. The average hourly compensation exceeds the minimum wage per hour in Russia. Each annotator is informed about topics that may be sensitive in the data, such as politics, societal minorities, and religion. 
Table \ref{tab:sets_create} summarizes general details concerning the creation of the datasets via crowd-source on ABC\footnote{\href{https://elementary.activebc.ru}{https://elementary.activebc.ru}} data labeling platform. 


\begin{table*}[ht!]
    \centering
    \begin{tabular}{@{}cp{0.20\linewidth}p{0.10\linewidth}p{0.10\linewidth}lp{0.11\linewidth}lp{0.11\linewidth}lp{0.1\linewidth}@{}}
        \toprule
         & \textbf{Task Name} & \textbf{Total} & \textbf{Pay Rate} & \textbf{Example Number} & \textbf{Overlap} \\
        \midrule
         & {\librusecHistory} & 84\$ &  6.25\$/hr & 32 & 1 \\
         & {\librusecMHQA} & 458\$&  6.25\$/hr & 40 & 3 \\
         & {\ruSciAbstractRetrieval} & 290\$ & 6.25\$/hr & 100 & 3 \\
        \bottomrule
    \end{tabular}
    \caption{The details of datasets collection.
    \textbf{Total} is the budget spent to annotate the tasks employed for metric evaluation.
    \textbf{Pay Rate} is the hourly rate computed as a simple average of pay rates based on time spent annotating one row and the reward for this row.
    \textbf{Example Number} refers to the total number of samples processed while collecting or verifying the dataset.
    \textbf{Overlap} is the median number of votes per dataset sample averaged across all annotation tasks for the same dataset (if more than 1 task is provided).
    }
    \label{tab:sets_create}
\end{table*}

\section{Dataset Examples}
\label{app:examples}
This section provides examples of the task format for the benchmark datasets. The exact prompts for the benchmark are not fixed. Here we provide prompts used in our experiments\footnote{All examples are presented in English for transparency and are given and are for illustrative purposes only to clarify the idea of a given task. The examples are not necessarily a direct translation of specific examples from the dataset. The exact prompts in their original formulation in Russian can be found in our repository \url{https://github.com/ai-forever/LIBRA}.}. 
\\\\
\noindent
\textbf{\passkey}: \emph{You are provided with a long text that contains the access key. Just remember the access key.}\\
Context: \{\emph{context}\}\\
\emph{You only need to specify the access key in the response.}\\
Question: \{\emph{input}\}\\
Answer:
\\\\
\noindent
\textbf{\passkeyWithoutGarbage}: \emph{You are provided with a long text that contains the access key. Just remember the access key.}\\
Context: \{\emph{context}\}\\
\emph{You only need to specify the access key in the response.}\\
Question: \{\emph{input}\}\\
Answer:
\\\\
\noindent
\textbf{\matreshkaNames}: \emph{You are provided with several dialogues. Remember the names of the people and the topics they talked about.}\\
Context: \{\emph{context}\}\\
\emph{In the answer, specify only the name of the interlocutor who spoke on the topic from the next question.}\\
Question: \{\emph{input}\}\\
Answer:
\\\\
\noindent
\textbf{\matreshkaYesNo}: \emph{You are provided with several dialogues. Remember the names of the topics that the interlocutors talked about.}\\
Context: \{\emph{context}\}\\
\emph{In the answer, you only need to specify 'Yes' if there was such a topic and 'No' if there was no such topic in the dialogues.}\\
Question: \{\emph{input}\}\\
Answer:
\\\\
\noindent
\textbf{\librusecHistory}: \emph{You are given a long text in which you need to find the answer to the question.}\\
Context: \{\emph{context}\}\\
\emph{Find the answer in the text to the following question.}\\
Question: \{\emph{input}\}\\
Answer:
\\\\
\noindent
\textbf{\ruTrec}: \emph{Define the type of question below. Here are some examples:}\\
Context: \{\emph{context}\}\\
\emph{Define the type of question below.}\\
Question: \{\emph{input}\}\\
Answer:
\\\\
\noindent
\textbf{\ruSciFi}: \emph{You are given a long text in which you need to find the answer to the question.}\\
Context: \{\emph{context}\}\\
\emph{You need to answer the following question with one of the options: 'False [in the real world: False]', 'True [in the real world: False]', 'True [in the real world: True]' or 'False [in the real world: True]'.}\\
Question: \{\emph{input}\}\\
Answer:
\\\\
\noindent
\textbf{\ruSciAbstractRetrieval}: \emph{Below are a few paragraphs. Determine which paragraph the short description corresponds to.}\\
Context: \{\emph{context}\}\\
\emph{Determine which paragraph the short description corresponds to. The response must contain the paragraph number.}\\
Question: \{\emph{input}\}\\
Answer:
\\\\
\noindent
\textbf{\ruTPO}: \emph{You are given a long text in which you need to find the answer to the question.}\\
Context: \{\emph{context}\}\\
\emph{You will be given several answers to the question in the text; choose only one correct one and specify the letter A, B, C, or D.}\\
Question: \{\emph{input}\}\\
Answer:
\\\\
\noindent
\textbf{\ruQuality}: \emph{You are given a long text in which you need to find the answer to the question.}\\
Context: \{\emph{context}\}\\
\emph{You will be given several answers to the question in the text; choose only one correct one.}\\
Question: \{\emph{input}\}\\
Answer:
\\\\
\noindent
\textbf{\LongContextMultiQ}: \emph{You are given a long text where you need to find the answer to the question.}\\
Context: \{\emph{context}\}\\
\emph{Find the answer in the text to the following question.}\\
Question: \{\emph{input}\}\\
Answer:
\\\\
\noindent
\textbf{\librusecMHQA}: \emph{You are given a long text where you need to find the answer.}\\
Context: \{\emph{context}\}\\
\emph{Find the answer in the text to the following question.}\\
Question: \{\emph{input}\}\\
Answer:
\\\\
\noindent
\textbf{\wikimqa}: \emph{The answer to the question is based on the above excerpts.}\\
Context: \{\emph{context}\}\\
\emph{Answer the question briefly, based on the above excerpts.}\\
Question: \{\emph{input}\}\\
Answer:
\\\\
\noindent
\textbf{\qaFirst}: \emph{I'm giving you a context with facts about the location of different people. You need to answer the question based only on information obtained from the facts. If the person was in different places, use the last location to answer the question.}\\
Context: \{\emph{context}\}\\
\emph{Answer the question as briefly as possible.}\\
Question: \{\emph{input}\}\\
Answer:
\\\\
\noindent
\textbf{\qaSecond}: \emph{I'm giving you a context with facts about the location and actions of different people. You need to answer the question based only on factual information. If a person took an item in one place and went to another, that item is also in the second place. If a person leaves an item in the first place and moves to the second place, the item remains in the first place.}\\
Context: \{\emph{context}\}\\
\emph{Answer the question as briefly as possible.}\\
Question: \{\emph{input}\}\\
Answer:
\\\\
\noindent
\textbf{\qaThird}: \emph{I'm giving you a context with facts about the location and actions of different people. You need to answer the question based only on factual information. If a person took an item in one place and went to another, that item is also in the second place. If a person leaves an item in the first mets and moves to the second place, the item remains in the first place.}\\
Context: \{\emph{context}\}\\
\emph{Answer the question as briefly as possible.}\\
Question: \{\emph{input}\}\\
Answer:
\\\\
\noindent
\textbf{\qaFourth}: \emph{I'm giving you a context with facts about the location and actions of different people. You need to answer the question based only on factual information.}\\
Context: \{\emph{context}\}\\
\emph{Answer the question as briefly as possible.}\\
Question: \{\emph{input}\}\\
Answer:
\\\\
\noindent
\textbf{\qaFifth}: \emph{I'm giving you a context with facts about the location and actions of different people. You need to answer the question based only on factual information.}\\
Context: \{\emph{context}\}\\
\emph{Answer the question as briefly as possible.}\\
Question: \{\emph{input}\}\\
Answer:
\\\\
\noindent
\textbf{\ruSciPassageCount}: \emph{Below are a few paragraphs. Read them and determine the number of unique paragraphs.}\\
Context: \{\emph{context}\}\\
\emph{Determine the number of unique paragraphs. The answer must contain only one number.}\\
Question: \{\emph{input}\}\\
Answer:
\\\\
\noindent
\textbf{\ruQasper}: \emph{You are provided with a scientific article and a question.}\\
Context: \{\emph{context}\}\\
\emph{Answer the question as briefly as possible, using a single phrase or sentence if possible. Don't give any explanations.}\\
Question: \{\emph{input}\}\\
Answer:
\\\\
\noindent
\textbf{\ruGsm}: \emph{Examples of mathematical problems are given below. Think step by step and answer the question.}\\
Context: \{\emph{context}\}\\
\emph{Think step by step and answer the question.}\\
Question: \{\emph{input}\}\\
Answer:

\section{Detailed Model Information}
\label{app:model}
The baseline model specifics are presented in Table~\ref{tab:model_details}. 

\begin{table*}
  \centering
  \small
  \renewcommand\arraystretch{0.95}
  \begin{tabular}{lccc}
    \toprule
    \textbf{Model Name} & \textbf{Type} & \textbf{Parameters}  & \textbf{Max Context Length} \\
    \midrule
        GPT-4o & Commercial & - & 128k\\
        GLM4-9B-Chat & Open-source & 9B & 128k\\
        Mistral-7B-Instruct-v0.3 & Open-source & 7B & 32k\\
        Mistral-7B-v0.3 & Open-source & 7B & 32k \\
        LLaMA-2-7B-32K & Open-source & 7B & 32k \\
        LongChat-7B-v1.5-32k & Open-source & 7B & 32k \\
        ChatGLM2-6B-32k & Open-source & 6B & 32k \\
        LongAlpaca-7B & Open-source & 7B & 32k\\
        LLaMA-3-8B-Instruct & Open-source & 8B & 8k\\
        Saiga-LLaMA-3-8B & Open-source & 8B & 8k\\
        LLaMA-3-8B & Open-source & 8B & 8k\\
        Mistral-7B-v0.1 & Open-source & 7B & 8k\\
    \bottomrule
  \end{tabular}
  \caption{The models evaluated as baselines. \textbf{Model Name} shows the name of the model.
    The \textbf{Max Context Length} shows maximal context lengths.
}
    \label{tab:model_details}
\end{table*}

\section{Detailed Model Results}
\label{app:results}
This section presents the detailed results of model evaluation. The results are shown for the following models:  GPT-4o (Table ~\ref{tab:result_chatgpt}), GLM4-9B-Chat (Table ~\ref{tab:result_glm4_9b}), Mistral-7B-Instruct-v0.3 (Table ~\ref{tab:tab:result_mistral_7b_v03_instr}), Mistral-7B-v0.3 (Table ~\ref{tab:result_mistral_7b_v03}), LLaMA-2-7B-32K (Table ~\ref{tab:result_llama2_32k}), LongChat-7B-v1.5-32k (Table ~\ref{tab:result_longchat32k}), ChatGLM2-6B-32K (Table ~\ref{tab:result_glm2_6b_32k}), LongAlpaca (Table ~\ref{tab:longalpaca}), LLaMA-3-8B-Instruct (Table ~\ref{tab:result_llama3_instr}), Saiga-LLaMA-3-8B (Table ~\ref{tab:result_saiga_llama3}), LLaMA-3-8B (Table ~\ref{tab:result_llama3}) and Mistral-7B-v0.1 (Table ~\ref{tab:result_mistral_7b_v01}).  

\begin{table*}
\small
  \centering
  \renewcommand\arraystretch{0.95}
  \begin{tabular}{@{}cp{0.20\linewidth}lccccccc@{}}
    \toprule
    & \textbf{Dataset Name} & \textbf{4k} & \textbf{8k} & \textbf{16k} & \textbf{32k} & \textbf{64k} & \textbf{128k} & \textbf{Overall} \\
    \midrule
    \multirow{2}{*}{\rotatebox[origin=c]{90}{\textbf{I}}}
    & {\passkey} & 100.0 & 100.0 & 100.0 & 100.0 & 100.0 & 100.0 & 100.0 \\
    & {\passkeyWithoutGarbage} & 100.0 & 100.0 & 100.0 & 100.0 & 100.0 & 100.0 & 100.0\\
    \midrule
    \multirow{8}{*}{\rotatebox[origin=c]{90}{\textbf{II}}}
        &{\matreshkaNames} & 60.0 & 60.0 & 50.0 & 40.0 & 50.0 & 50.0 & 51.7  \\
        &{\matreshkaYesNo} & 80.0 & 60.0 & 100.0 & 80.0 & 70.0 & 90.0 & 80.0 \\
        &{\librusecHistory} & - & 100.0 & 100.0 & 100.0 & 90.0 & - & 97.5\\
        &{\ruTrec} & 60.0 & 80.0 & 90.0 & 70.0 & - & - & 75.0\\
        &{\ruSciFi} & - & - & - & 60.0 & 90.0 & - & 75.0\\
        &{\ruSciAbstractRetrieval} & 99.0 & 95.4 & 92.5 & 95.6 & 59.1 & 19.8 & 76.9  \\
        &{\ruTPO} & - & 100.0 & - & - & - & - & 100.0 \\
        &{\ruQuality} & - & 80.0 & 86.7 & - & - &  -& 83.3\\
    \midrule
    \multirow{8}{*}{\rotatebox[origin=c]{90}{\textbf{III}}}
    &{\LongContextMultiQ} & 30.0 & 100.0 & 70.0 & 0.0 & 10.0 & 10.0 & 36.7\\
    &{\librusecMHQA} & - & 50.0 & - & - & - & - & 50.0 \\
    &{\wikimqa} & - & 80.0 & 80.0 & 70.0 & -  & - & 76.7 \\
    &{\qaFirst} &  90.0 & 80.0 & 70.0 & 90.0 & 80.0 & 60.0 & 78.3 \\
    &{\qaSecond} & 40.0 & 30.0 & 40.0 & 40.0 & 50.0 & 20.0 & 36.7 \\
    &{\qaThird} & 20.0 & 30.0 & 10.0 & 20.0 & 20.0 & 28.7 & 21.4 \\
    &{\qaFourth} & 88.0 & 80.0 & 80.0 & 57.1 & 88.6 & 80.0 & 79.0 \\
    &{\qaFifth} & 86.7 & 86.7 & 93.3 & 96.7 & 86.7 & 90.0 & 90.0 \\
    \midrule
    \multirow{3}{*}{\rotatebox[origin=c]{90}{\textbf{IV}}}
    &{\ruSciPassageCount} & 100.0 & 50.0 & 30.0 & 0.0 & 20.0 & 10.0 & 35.0\\
    &{\ruQasper} & - & 28.7 & 31.8 & 34.7 & - & - & 31.7  \\
    &{\ruGsm}  & - & - & 100.0 & - & - & - & 100.0  \\
    \bottomrule
    \end{tabular}
  \caption{The table presents the evaluation results of GPT-4o. \textbf{Dataset Name} shows the name of the dataset.
    The rows \textbf{4k}, \textbf{8k}, \textbf{16k}, \textbf{32k}, \textbf{64k}, \textbf{128k} show evaluation scores of datasets for each context length, respectively. The \textbf{Overall} score is obtained by averaging the results over each length.
} \label{tab:result_chatgpt}
\end{table*}
\begin{table*}
\small
  \centering
  \renewcommand\arraystretch{0.95}
  \begin{tabular}{@{}cp{0.20\linewidth}lccccccc@{}}
    \toprule
    &\textbf{Dataset Name} & \textbf{4k} & \textbf{8k} & \textbf{16k} & \textbf{32k} & \textbf{64k} & \textbf{128k} & \textbf{Overall} \\
    \midrule
    \multirow{2}{*}{\rotatebox[origin=c]{90}{\textbf{I}}}
    &{\passkey} & 100.0 & 100.0 & 100.0 & 100.0 & 100.0 & 100.0 & 100.0 \\
    &{\passkeyWithoutGarbage} & 100.0 & 100.0 & 100.0 & 100.0 & 100.0 & 100.0 & 100.0\\
    \midrule
    \multirow{8}{*}{\rotatebox[origin=c]{90}{\textbf{II}}}
    &{\matreshkaNames} & 64.7 & 50.7 & 52.0 & 47.3 & 37.3 & 32.0 & 47.3  \\
    &{\matreshkaYesNo} & 79.3 & 75.0 & 71.3 & 67.0 & 59.7 & 56.0 & 68.0 \\
    &{\librusecHistory} & - & 84.4 & 84.4 & 84.4 & 75.0 & - & 82.0 \\
    &{\ruTrec} & 56.8 & 70.0 & 75.8 & 77.0 & - & - & 69.9\\
    &{\ruSciFi} & - & - & - & 38.9 & 42.9 & - & 40.9\\
    &{\ruSciAbstractRetrieval} & 98.2 & 92.3 & 91.2 & 81.9 & 64.1 & 39.1 & 77.8  \\
    &{\ruTPO} & - & 86.9 & - & - & - &  -& 86.9 \\
    &{\ruQuality} &-  & 82.9 & 65.2 & - & - & - & 74.1\\
    \midrule
    \multirow{8}{*}{\rotatebox[origin=c]{90}{\textbf{III}}}
    &{\LongContextMultiQ} & 5.5 & 26.5 & 3.5 & 0.5 & 0.5 & 10.0 & 7.8\\
    &{\librusecMHQA} & - & 44.5 & - & - & - & - & 44.5 \\
    &{\wikimqa} &  - & 55.1 & 55.5 & 35.8 & - & - & 48.8  \\
    &{\qaFirst} &  69.9 & 59.0 & 60.0 & 50.8 & 42.9 & 42.0 & 54.1 \\
    &{\qaSecond} & 38.9 & 33.0 & 29.9 & 26.9 & 26.8 & 23.5 & 29.8 \\
    &{\qaThird} & 24.6 & 27.9 & 21.4 & 22.6 & 18.7 & 18.5 & 22.3 \\
    &{\qaFourth} & 62.1 & 59.6 & 56.6 & 58.0 & 43.0 & 37.7 & 52.8 \\
    &{\qaFifth} & 73.0 & 73.5 & 72.0 & 66.8 & 69.7 & 67.0 & 70.3 \\
    \midrule
    \multirow{3}{*}{\rotatebox[origin=c]{90}{\textbf{IV}}}
    &{\ruSciPassageCount} & 27.0 & 8.0 & 9.0 & 0.0 & 1.0 & 0.0 & 7.5\\
    &{\ruQasper} & - & 6.5 & 5.9 & 2.6 & - & - & 5.0 \\
    &{\ruGsm}  & - & - & 8.0 & - & - & - & 8.0 \\
    \bottomrule
    \end{tabular}
  \caption{The table presents the evaluation results of GLM4-9B. \textbf{Dataset Name} shows the name of the dataset.
    The rows \textbf{4k}, \textbf{8k}, \textbf{16k}, \textbf{32k}, \textbf{64k}, \textbf{128k} show evaluation scores of datasets for each context length, respectively. The \textbf{Overall} score is obtained by averaging the results over each length.
} \label{tab:result_glm4_9b}
\end{table*}
\begin{table*}
\small
  \centering
  \renewcommand\arraystretch{0.95}
\begin{tabular}{@{}cp{0.20\linewidth}lccccccc@{}}
    \toprule
    &\textbf{Dataset Name} & \textbf{4k} & \textbf{8k} & \textbf{16k} & \textbf{32k} & \textbf{64k} & \textbf{128k} & \textbf{Overall} \\
    \midrule
    \multirow{2}{*}{\rotatebox[origin=c]{90}{\textbf{I}}}
    &{\passkey} & 100.0 & 100.0 & 100.0 & 100.0 & 0.0 & 0.0 & 66.7 \\
    &{\passkeyWithoutGarbage} & 100.0 & 100.0 & 100.0 & 99.5 & 0.0 & 0.0 & 66.6 \\
    \midrule
    \multirow{8}{*}{\rotatebox[origin=c]{90}{\textbf{II}}}
    &{\matreshkaNames} & 38.0 & 32.0 & 16.7 & 11.3 & 0.0 & 0.0 & 16.3 \\
    &{\matreshkaYesNo} & 56.5 & 50.7 & 54.7 & 50.0 & 0.0 & 0.0 & 35.3\\
    &{\librusecHistory} & - & 71.9 & 62.5 & 68.8 & 0.0 & - & 50.8 \\
    &{\ruTrec} &  56.8 & 38.0 & 40.7 & 34.4 & - & - & 42.5 \\
    &{\ruSciFi} & - & - & - & 30.6 & 0.0 & - & 15.3\\
    &{\ruSciAbstractRetrieval} & 98.2 & 86.9 & 71.1 & 5.1 & 0.0 & 0.0 & 43.6 \\
    &{\ruTPO} & - & 66.4 & - & - & - & - & 66.4\\
    &{\ruQuality} & - & 38.2 & 23.0 & - & - & - & 30.6 \\
    \midrule
    \multirow{8}{*}{\rotatebox[origin=c]{90}{\textbf{III}}}
    &{\LongContextMultiQ} & 3.5 & 22.0 & 3.5 & 0.0 & 0.0 & 0.0 & 4.8 \\
    &{\librusecMHQA} & - & 33.6 & - & - & - & - & 33.6\\
    &{\wikimqa} & - & 55.1 & 46.9 & 27.6 & - & - & 43.2 \\
    &{\qaFirst} & 25.0 & 15.0 & 22.0 & 24.0 & 0.0 & 0.0 & 14.3 \\
    &{\qaSecond} & 8.0 & 5.0 & 2.0 & 2.0 & 0.0 & 0.0 & 2.8  \\
    &{\qaThird} &  10.0 & 8.0 & 10.0 & 8.0 & 0.0 & 0.0 & 6.0 \\
    &{\qaFourth} & 51.8 & 44.3 & 39.3 & 30.3 & 0.0 & 0.0 & 27.6 \\
    &{\qaFifth} & 54.7 & 62.0 & 55.3 & 53.3 & 0.0 & 0.0 & 37.6 \\
    \midrule
    \multirow{3}{*}{\rotatebox[origin=c]{90}{\textbf{IV}}}
    &{\ruSciPassageCount} & 26.0 & 14.0 & 7.0 & 2.0 & 0.0 & 0.0 & 8.2 \\
    &{\ruQasper} & - & 6.6 & 6.6 & 2.9 & - & - & 5.4\\
    &{\ruGsm}  & - & - & 11.0 & - & - & - & 11.0 \\
    \bottomrule
    \end{tabular}
  \caption{The table presents the evaluation results of Mistral-7B-v0.3-Instruct. \textbf{Dataset Name} shows the name of the dataset.
    The rows \textbf{4k}, \textbf{8k}, \textbf{16k}, \textbf{32k}, \textbf{64k}, \textbf{128k} show evaluation scores of datasets for each context length, respectively. The \textbf{Overall} score is obtained by averaging the results over each length.
} \label{tab:tab:result_mistral_7b_v03_instr}
\end{table*}
\begin{table*}
\small
  \centering
  \renewcommand\arraystretch{0.95}
\begin{tabular}{@{}cp{0.20\linewidth}lccccccc@{}}
    \toprule
    &\textbf{Dataset Name} & \textbf{4k} & \textbf{8k} & \textbf{16k} & \textbf{32k} & \textbf{64k} & \textbf{128k} & \textbf{Overall} \\
    \midrule
    \multirow{2}{*}{\rotatebox[origin=c]{90}{\textbf{I}}}
    &{\passkey} & 100.0 & 100.0 & 100.0 & 100.0 & 0.0 & 0.0 & 66.7 \\
    &{\passkeyWithoutGarbage} & 100.0 & 100.0 & 100.0 & 100.0 & 0.0 & 0.0 & 66.7 \\
    \midrule
    \multirow{8}{*}{\rotatebox[origin=c]{90}{\textbf{II}}}
    &{\matreshkaNames} & 28.7 & 16.0 & 10.7 & 4.7 & 0.0 & 0.0 & 10.0 \\
    &{\matreshkaYesNo} & 44.8 & 47.0 & 50.0 & 50.0 & 0.0 & 0.0 & 32.0\\
    &{\librusecHistory} & - & 93.8 & 93.8 & 84.4 & 0.0 & - & 68.0 \\
    &{\ruTrec} &  0.0 & 8.0 & 4.4 & 9.0 & - & - & 5.4 \\
    &{\ruSciFi} & - & - & - & 0.0 & 0.0 & - & 0.0\\
    &{\ruSciAbstractRetrieval} & 87.4 & 56.6 & 36.9 & 1.9 & 0.0 & 0.0 & 30.5\\
    &{\ruTPO} & - & 39.7 & - & - & - & - & 39.7\\
    &{\ruQuality} & - & 23.6 & 6.8 & - & - & - & 15.2\\
    \midrule
    \multirow{8}{*}{\rotatebox[origin=c]{90}{\textbf{III}}}
    &{\LongContextMultiQ} & 4.0 & 24.0 & 3.5 & 0.0 & 0.0 & 0.0 & 5.2 \\
    &{\librusecMHQA} & - & 39.1 &-  & - & - & - & 39.1 \\
    &{\wikimqa} & -  & 46.9 & 49.2 & 26.8 & - & - & 41.0\\
    &{\qaFirst} &  60.0 & 63.0 & 58.0 & 43.0 & 0.0 & 0.0 & 37.3 \\
    &{\qaSecond} & 35.0 & 23.0 & 18.0 & 24.0 & 0.0 & 0.0 & 16.7  \\
    &{\qaThird} &  29.0 & 23.0 & 23.0 & 19.0 & 0.0 & 0.0 & 15.7 \\
    &{\qaFourth} & 46.3 & 34.4 & 36.2 & 24.9 & 0.0 & 0.0 & 23.6 \\
    &{\qaFifth} & 70.3 & 68.7 & 75.3 & 68.3 & 0.0 & 0.0 & 47.1 \\
    \midrule
    \multirow{3}{*}{\rotatebox[origin=c]{90}{\textbf{IV}}}
    &{\ruSciPassageCount} & 0.0 & 0.0 & 0.0 & 0.0 & 0.0 & 0.0 & 0.0 \\
    &{\ruQasper} & - & 8.9 & 6.5 & 1.9 & - & - & 5.8\\
    &{\ruGsm}  & - & - & 9.0 & - & - & - & 9.0 \\
    \bottomrule
    \end{tabular}
  \caption{The table presents the evaluation results of Mistral-7B-v0.3. \textbf{Dataset Name} shows the name of the dataset.
    The rows \textbf{4k}, \textbf{8k}, \textbf{16k}, \textbf{32k}, \textbf{64k}, \textbf{128k} show evaluation scores of datasets for each context length, respectively. The \textbf{Overall} score is obtained by averaging the results over each length.
} \label{tab:result_mistral_7b_v03}
\end{table*}
\begin{table*}
\small
  \centering
  \renewcommand\arraystretch{0.95}
\begin{tabular}{@{}cp{0.20\linewidth}lccccccc@{}}
    \toprule
    &\textbf{Dataset Name} & \textbf{4k} & \textbf{8k} & \textbf{16k} & \textbf{32k} & \textbf{64k} & \textbf{128k} & \textbf{Overall} \\
    \midrule
    \multirow{2}{*}{\rotatebox[origin=c]{90}{\textbf{I}}}
    &{\passkey} &  100.0 & 100.0 & 100.0 & 100.0 & 0.0 & 0.0 & 66.7 \\
    &{\passkeyWithoutGarbage} & 100.0 & 97.5 & 98.5 & 97.0 & 0.0 & 0.0 & 65.5 \\
    \midrule
    \multirow{8}{*}{\rotatebox[origin=c]{90}{\textbf{II}}}
    &{\matreshkaNames} &  8.0 & 6.7 & 2.0 & 4.0 & 0.0 & 0.0 & 3.4  \\
    &{\matreshkaYesNo} & 50.2 & 50.0 & 50.0 & 50.0 & 0.0 & 0.0 & 33.4 \\
    &{\librusecHistory} & - &68.8 & 50.0 & 43.8 & 0.0 & - & 40.6 \\
    &{\ruTrec} & 24.3 & 18.0 & 24.2 & 28.7 & - & - & 23.8\\
    &{\ruSciFi} & - & - & - & 11.1 & 0.0 & - & 5.6 \\
    &{\ruSciAbstractRetrieval} & 85.2 & 76.1 & 46.8 & 26.7 & 0.0 & 0.0 & 39.1 \\
    &{\ruTPO} & - & 54.3 & - & - & - & - & 54.3  \\
    &{\ruQuality} & - & 17.1 & 13.9 & - & - & - & 15.5 \\
    \midrule
    \multirow{8}{*}{\rotatebox[origin=c]{90}{\textbf{III}}}
    &{\LongContextMultiQ} & 4.5 & 33.0 & 10.0 & 0.0 & 0.0 & 0.0 & 7.9 \\
    &{\librusecMHQA} & - & 27.6 & -  & - & - & - & 27.6 \\
    &{\wikimqa} & - & 44.9 & 39.8 & 26.8 & - & - & 37.2  \\
    &{\qaFirst} & 60.0 & 66.0 & 66.0 & 50.0 & 0.0 & 0.0 & 40.3 \\
    &{\qaSecond} & 25.0 & 30.0 & 25.9 & 19.0 & 0.0 & 0.0 & 16.6 \\
    &{\qaThird} & 22.9 & 28.9 & 26.0 & 20.0 & 0.0 & 0.0 & 16.3 \\
    &{\qaFourth} & 31.0 & 34.0 & 23.0 & 12.0 & 0.0 & 0.0 & 16.7 \\
    &{\qaFifth} & 59.0 & 66.0 & 64.0 & 69.0 & 0.0 & 0.0 & 43.0 \\
    \midrule
    \multirow{3}{*}{\rotatebox[origin=c]{90}{\textbf{IV}}}
    &{\ruSciPassageCount} & 18.0 & 5.0 & 5.0 & 0.5 & 0.0 & 0.0 & 4.7\\
    &{\ruQasper} & - & 5.8 & 6.0 & 2.2 & - & - & 4.7\\
    &{\ruGsm}  &  - & - & 7.0 & - & - & - & 7.0 \\
    \bottomrule
    \end{tabular}
  \caption{The table presents the evaluation results of LLaMA-2-32K. \textbf{Dataset Name} shows the name of the dataset.
    The rows \textbf{4k}, \textbf{8k}, \textbf{16k}, \textbf{32k}, \textbf{64k}, \textbf{128k} show evaluation scores of datasets for each context length, respectively. The \textbf{Overall} score is obtained by averaging the results over each length.
} \label{tab:result_llama2_32k}
\end{table*}
\begin{table*}
\small
  \centering
  \renewcommand\arraystretch{0.95}
\begin{tabular}{@{}cp{0.20\linewidth}lccccccc@{}}
    \toprule
    &\textbf{Dataset Name} & \textbf{4k} & \textbf{8k} & \textbf{16k} & \textbf{32k} & \textbf{64k} & \textbf{128k} & \textbf{Overall} \\
    \midrule
    \multirow{2}{*}{\rotatebox[origin=c]{90}{\textbf{I}}}
    &{\passkey} & 99.5 & 100.0 & 100.0 & 99.5 & 0.0 & 0.0 & 66.5 \\
    &{\passkeyWithoutGarbage} & 100.0 & 100.0 & 98.5 & 97.5 & 0.0 & 0.0 & 66.0 \\
    \midrule
    \multirow{8}{*}{\rotatebox[origin=c]{90}{\textbf{II}}}
    &{\matreshkaNames} & 17.3 & 6.7 & 8.0 & 3.3 & 0.0 & 0.0 & 5.9 \\
    &{\matreshkaYesNo} & 50.2 & 50.0 & 50.0 & 50.0 & 0.0 & 0.0 & 33.4 \\
    &{\librusecHistory} & - & 56.2 & 34.4 & 15.6 & 0.0 & - & 26.6 \\
    &{\ruTrec} & 5.4 & 10.0 & 7.7 & 6.6 & - & - & 7.4 \\
    &{\ruSciFi} & -  & - & - & 5.6 & 0.0 & - & 2.8 \\
    &{\ruSciAbstractRetrieval} & 87.4 & 76.2 & 60.6 & 22.2 & 0.0 & 0.0 & 41.1 \\
    &{\ruTPO} & - & 39.6 & - & - & - & - & 39.6\\
    &{\ruQuality} &  - & 28.5 & 17.8 & - & - & - & 23.1 \\
    \midrule
    \multirow{8}{*}{\rotatebox[origin=c]{90}{\textbf{III}}}
    &{\LongContextMultiQ} & 2.5 & 14.0 & 2.5 & 0.0 & 0.0 & 0.0 & 3.2 \\
    &{\librusecMHQA} & - & 24.7 & - & - & - & - & 24.7\\
    &{\wikimqa} & - & 42.9 & 39.8 & 22.8 & - & - & 35.2\\
    &{\qaFirst} & 26.0 & 29.0 & 31.0 & 19.0 & 0.0 & 0.0 & 17.5 \\
    &{\qaSecond} & 11.0 & 8.0 & 16.0 & 8.0 & 0.0 & 0.0 & 7.2 \\
    &{\qaThird} &  9.0 & 5.0 & 4.0 & 6.0 & 0.0 & 0.0 & 4.0 \\
    &{\qaFourth} & 25.2 & 29.2 & 15.6 & 5.9 & 0.0 & 0.0 & 12.7 \\
    &{\qaFifth} & 51.3 & 50.0 & 48.3 & 50.0 & 0.0 & 0.0 & 33.3 \\
    \midrule
    \multirow{3}{*}{\rotatebox[origin=c]{90}{\textbf{IV}}}
    &{\ruSciPassageCount} &18.0 & 8.0 & 1.0 & 2.0 & 0.0 & 0.0 & 4.8 \\
    &{\ruQasper} & - & 6.1 & 6.5 & 2.4 & - & - & 5.0\\
    &{\ruGsm}  & - & - & 5.0 & - & - & - & 5.0  \\
    \bottomrule
    \end{tabular}
  \caption{The table presents the evaluation results of LongChat. \textbf{Dataset Name} shows the name of the dataset.
    The rows \textbf{4k}, \textbf{8k}, \textbf{16k}, \textbf{32k}, \textbf{64k}, \textbf{128k} show evaluation scores of datasets for each context length, respectively. The \textbf{Overall} score is obtained by averaging the results over each length.
} \label{tab:result_longchat32k}
\end{table*}
\begin{table*}
\small
  \centering
  \renewcommand\arraystretch{0.95}
  \begin{tabular}{@{}cp{0.20\linewidth}lccccccc@{}}
    \toprule
    &\textbf{Dataset Name} & \textbf{4k} & \textbf{8k} & \textbf{16k} & \textbf{32k} & \textbf{64k} & \textbf{128k} & \textbf{Overall} \\
    \midrule
    \multirow{2}{*}{\rotatebox[origin=c]{90}{\textbf{I}}}
    &{\passkey} & 100.0 & 100.0 & 100.0 & 82.0 & 0.0 & 0.0 & 63.7 \\
    &{\passkeyWithoutGarbage} & 99.0 & 99.5 & 98.5 & 93.0 & 0.0 & 0.0 & 65.0 \\
    \midrule
    \multirow{8}{*}{\rotatebox[origin=c]{90}{\textbf{II}}}
    &{\matreshkaNames} & 4.7 & 2.7 & 0.7 & 0.0 & 0.0 & 0.0 & 1.3 \\
    &{\matreshkaYesNo} & 50.2 & 50.0 & 50.0 & 50.0 & 0.0 & 0.0 & 33.4 \\
    &{\librusecHistory} & - & 21.9 & 9.4 & 3.1 & 0.0 & - & 8.6 \\
    &{\ruTrec} & 5.4 & 4.0 & 4.4 & 4.1 & - & - & 4.5 \\
    &{\ruSciFi} & - & - & - & 0.0 & 0.0 & - & 0.0 \\
    &{\ruSciAbstractRetrieval} & 41.7 & 21.9 & 18.1 & 0.0 & 0.0 & 0.0 & 13.6 \\
    &{\ruTPO} & - & 29.0 & - & - & - & - & 29.0 \\
    &{\ruQuality} & - & 54.5 & 43.9 & - & - & - & 49.2 \\
    \midrule
    \multirow{8}{*}{\rotatebox[origin=c]{90}{\textbf{III}}}
    &{\LongContextMultiQ} & 0.5 & 5.0 & 1.5 & 0.0 & 0.0 & 0.0 & 1.2 \\
    &{\librusecMHQA} & - & 6.8 & - & - & - & - & 6.8 \\
    &{\wikimqa} & - & 18.4 & 21.9 & 12.2 & - & - & 17.5 \\
    &{\qaFirst} & 27.0 & 23.0 & 23.0 & 0.0 & 0.0 & 0.0 & 12.2 \\
    &{\qaSecond} & 5.0 & 4.0 & 0.0 & 0.0 & 0.0 & 0.0 & 1.5 \\
    &{\qaThird} & 7.0 & 3.0 & 4.0 & 1.0 & 0.0 & 0.0 & 2.5 \\
    &{\qaFourth} & 2.0 & 1.8 & 0.0 & 0.0 & 0.0 & 0.0 & 0.6 \\
    &{\qaFifth} & 20.0 & 18.0 & 15.0 & 0.0 & 0.0 & 0.0 & 8.8 \\
    \midrule
    \multirow{3}{*}{\rotatebox[origin=c]{90}{\textbf{IV}}}
    &{\ruSciPassageCount} & 9.0 & 6.0 & 7.0 & 0.0 & 0.0 & 0.0 & 3.7 \\
    &{\ruQasper} & - & 3.5 & 3.3 & 0.9 & - & - & 2.6 \\
    &{\ruGsm}  & - & - & 5.0 & - & - & - & 5.0 \\
    \bottomrule
    \end{tabular}
  \caption{The table presents the evaluation results of GLM2-6B-32K. \textbf{Dataset Name} shows the name of the dataset.
    The rows \textbf{4k}, \textbf{8k}, \textbf{16k}, \textbf{32k}, \textbf{64k}, \textbf{128k} show evaluation scores of datasets for each context length, respectively. The \textbf{Overall} score is obtained by averaging the results over each length.
} \label{tab:result_glm2_6b_32k}
\end{table*}
\begin{table*}
\small
  \centering
  \renewcommand\arraystretch{0.95}
\begin{tabular}{@{}cp{0.20\linewidth}lccccccc@{}}
    \toprule
    &\textbf{Dataset Name} & \textbf{4k} & \textbf{8k} & \textbf{16k} & \textbf{32k} & \textbf{64k} & \textbf{128k} & \textbf{Overall} \\
    \midrule
    \multirow{2}{*}{\rotatebox[origin=c]{90}{\textbf{I}}}
    & {\passkey} & 77.5 & 82.5 & 57.5 & 37.0 & 0.0 & 0.0 & 42.4 \\
    &{\passkeyWithoutGarbage} & 71.0 & 70.0 & 56.0 & 46.5 & 0.0 & 0.0 & 40.6 \\
    \midrule
    \multirow{8}{*}{\rotatebox[origin=c]{90}{\textbf{II}}}
    &{\matreshkaNames} & 1.3 & 0.7 & 0.7 & 0.0 & 0.0 & 0.0 & 0.4 \\
    &{\matreshkaYesNo} & 47.8 & 39.3 & 48.0 & 47.7 & 0.0 & 0.0 & 30.5 \\
    &{\librusecHistory} & - & 18.8 & 15.6 & 18.8 & 0.0 & - & 13.3 \\
    &{\ruTrec} & 0.0 & 2.0 & 0.0 & 0.0 & - & - & 0.5 \\
    &{\ruSciFi} & - & - & - & 2.8 & 0.0 & - & 1.4 \\
    &{\ruSciAbstractRetrieval} & 65.0 & 44.7 & 20.4 & 11.2 & 0.0 & 0.0 & 23.5 \\
    &{\ruTPO} & - & 6.8 & - &  - & - & - & 6.8 \\
    &{\ruQuality} & - & 39.8 & 48.2 & - & - & - & 44.0 \\
    \midrule
    \multirow{8}{*}{\rotatebox[origin=c]{90}{\textbf{III}}}
    &{\LongContextMultiQ} & 3.0 & 1.5 & 0.0 & 0.0 & 0.0 & 0.0 & 0.8 \\
    &{\librusecMHQA} & - & 7.8 & - & - & - & - & 7.8\\
    &{\wikimqa} & - & 40.8 & 28.9 & 21.1 &  - & - & 30.3\\
    &{\qaFirst} & 9.0 & 6.0 & 6.0 & 2.0 & 0.0 & 0.0 & 3.8 \\
    &{\qaSecond} & 1.0 & 1.0 & 0.0 & 0.0 & 0.0 & 0.0 & 0.3 \\
    &{\qaThird} &  5.0 & 9.0 & 4.0 & 2.9 & 0.0 & 0.0 & 3.5 \\
    &{\qaFourth} & 0.0 & 1.0 & 0.0 & 0.0 & 0.0 & 0.0 & 0.2 \\
    &{\qaFifth} & 44.2 & 44.0 & 47.5 & 40.7 & 0.0 & 0.0 & 29.4 \\
    \midrule
    \multirow{3}{*}{\rotatebox[origin=c]{90}{\textbf{IV}}}
    &{\ruSciPassageCount} & 13.0 & 5.0 & 2.0 & 3.0 & 0.0 & 0.0 & 3.8 \\
    &{\ruQasper} & - & 2.3 & 2.2 & 1.6 & - & - & 2.0\\
    &{\ruGsm}  & - & - & 2.0 & - & - & - & 2.0  \\
    \bottomrule
    \end{tabular}
  \caption{The table presents the evaluation results of LongAlpaca. \textbf{Dataset Name} shows the name of the dataset.
    The rows \textbf{4k}, \textbf{8k}, \textbf{16k}, \textbf{32k}, \textbf{64k}, \textbf{128k} show evaluation scores of datasets for each context length, respectively. The \textbf{Overall} score is obtained by averaging the results over each length.
} \label{tab:longalpaca}
\end{table*}
\begin{table*}
\small
  \centering
  \renewcommand\arraystretch{0.95}
 \begin{tabular}{@{}cp{0.20\linewidth}lccccccc@{}}
    \toprule
    & \textbf{Dataset Name} & \textbf{4k} & \textbf{8k} & \textbf{16k} & \textbf{32k} & \textbf{64k} & \textbf{128k} & \textbf{Overall} \\
    \midrule
    \multirow{2}{*}{\rotatebox[origin=c]{90}{\textbf{I}}}
    &{\passkey} & 100.0 & 100.0 & 0.0 & 0.0 & 0.0 & 0.0 & 33.3 \\
    &{\passkeyWithoutGarbage} & 100.0 & 100.0 & 0.0 & 0.0 & 0.0 & 0.0 & 33.3 \\
    \midrule
    \multirow{8}{*}{\rotatebox[origin=c]{90}{\textbf{II}}}
    &{\matreshkaNames} & 53.3 & 46.0 & 0.0 & 0.0 & 0.0 & 0.0 & 16.6 \\
    &{\matreshkaYesNo} & 83.9 & 80.0 & 0.0 & 0.0 & 0.0 & 0.0 & 27.3 \\
    &{\librusecHistory} & - & 90.6 & 0.0 & 0.0 & 0.0 & - & 22.7 \\
    &{\ruTrec} & 59.5 & 50.0 & 0.0 & 0.0 & - & - & 27.4 \\
    &{\ruSciFi} & - & - & - & 0.0 & 0.0 & - & 0.0 \\
    &{\ruSciAbstractRetrieval} & 96.6 & 91.5 & 0.0 & 0.0 & 0.0 & 0.0 & 31.4 \\
    &{\ruTPO} & - & 78.1 & - & - & - & - & 78.1 \\
    &{\ruQuality} & - & 69.1 & 0.0 & - & - & - & 34.6 \\
    \midrule
    \multirow{8}{*}{\rotatebox[origin=c]{90}{\textbf{III}}}
    &{\LongContextMultiQ} & 5.0 & 24.5 & 0.0 & 0.0 & 0.0 & 0.0 & 4.9 \\
    &{\librusecMHQA} & - & 46.1 & - & - & - & - & 46.1 \\
    &{\wikimqa} & - & 53.1 & 0.0 & 0.0 & - & - & 17.7 \\
    &{\qaFirst} & 68.6 & 73.4 & 0.0 & 0.0 & 0.0 & 0.0 & 23.7 \\
    &{\qaSecond} & 14.0 & 10.9 & 0.0 & 0.0 & 0.0 & 0.0 & 4.1 \\
    &{\qaThird} &  9.0 & 18.0 & 0.0 & 0.0 & 0.0 & 0.0 & 4.5 \\
    &{\qaFourth} & 57.3 & 60.3 & 0.0 & 0.0 & 0.0 & 0.0 & 19.6 \\
    &{\qaFifth} & 76.7 & 75.2 & 0.0 & 0.0 & 0.0 & 0.0 & 25.3 \\
    \midrule
    \multirow{3}{*}{\rotatebox[origin=c]{90}{\textbf{IV}}}
    &{\ruSciPassageCount} & 31.0 & 8.0 & 0.0 & 0.0 & 0.0 & 0.0 & 6.5 \\
    &{\ruQasper} & - & 6.5 & 0.0 & 0.0 & - & - & 2.2 \\
    &{\ruGsm}  & - & - & 0.0 & - & - & - & 0.0  \\
    \bottomrule
    \end{tabular}
  \caption{The table presents the evaluation results of LLaMA-3-8B-Instruct. \textbf{Dataset Name} shows the name of the dataset.
    The rows \textbf{4k}, \textbf{8k}, \textbf{16k}, \textbf{32k}, \textbf{64k}, \textbf{128k} show evaluation scores of datasets for each context length, respectively. The \textbf{Overall} score is obtained by averaging the results over each length.
} \label{tab:result_llama3_instr}
\end{table*}
\begin{table*}
\small
  \centering
  \renewcommand\arraystretch{0.95}
\begin{tabular}{@{}cp{0.20\linewidth}lccccccc@{}}
    \toprule
    &\textbf{Dataset Name} & \textbf{4k} & \textbf{8k} & \textbf{16k} & \textbf{32k} & \textbf{64k} & \textbf{128k} & \textbf{Overall} \\
    \midrule
    \multirow{2}{*}{\rotatebox[origin=c]{90}{\textbf{I}}}
    &{\passkey} & 100.0 & 100.0 & 0.0 & 0.0 & 0.0 & 0.0 & 33.3 \\
    &{\passkeyWithoutGarbage} & 100.0 & 99.5 & 0.0 & 0.0 & 0.0 & 0.0 & 33.2 \\
    \midrule
    \multirow{8}{*}{\rotatebox[origin=c]{90}{\textbf{II}}}
    &{\matreshkaNames} & 53.3 & 40.0 & 0.0 & 0.0 & 0.0 & 0.0 & 15.6 \\
    &{\matreshkaYesNo} & 87.3 & 81.0 & 0.0 & 0.0 & 0.0 & 0.0 & 28.0\\
    &{\librusecHistory} & - & 96.9 & 0.0 & 0.0 & 0.0 & - & 24.2 \\
    &{\ruTrec} & 51.4 & 54.0 & 0.0 & 0.0 & - & - & 26.3 \\
    &{\ruSciFi} & - & - & - & 0.0 & 0.0 & - & 0.0\\
    &{\ruSciAbstractRetrieval} & 97.7 & 92.6 & 0.0 & 0.0 & 0.0 & 0.0 & 31.7\\
    &{\ruTPO} & - & 75.7 & - & - & - & - & 75.7\\
    &{\ruQuality} & - & 35.8 & 0.0 & - & - & - & 17.9\\
    \midrule
    \multirow{8}{*}{\rotatebox[origin=c]{90}{\textbf{III}}}
    &{\LongContextMultiQ} & 5.5 & 23.5 & 0.0 & 0.0 & 0.0 & 0.0 & 4.8 \\
    &{\librusecMHQA} & -  & 45.1 & - & - & - & - & 45.1 \\
    &{\wikimqa} &-  & 53.1 & 0.0 & 0.0 & - & - & 17.7\\
    &{\qaFirst} &  76.3 & 75.8 & 0.0 & 0.0 & 0.0 & 0.0 & 25.4 \\
    &{\qaSecond} & 19.6 & 6.9 & 0.0 & 0.0 & 0.0 & 0.0 & 4.4  \\
    &{\qaThird} &  14.7 & 21.6 & 0.0 & 0.0 & 0.0 & 0.0 & 6.1 \\
    &{\qaFourth} & 63.5 & 58.2 & 0.0 & 0.0 & 0.0 & 0.0 & 20.3 \\
    &{\qaFifth} & 74.7 & 76.3 & 0.0 & 0.0 & 0.0 & 0.0 & 25.2 \\
    \midrule
    \multirow{3}{*}{\rotatebox[origin=c]{90}{\textbf{IV}}}
    &{\ruSciPassageCount} & 19.5 & 3.5 & 0.0 & 0.0 & 0.0 & 0.0 & 3.8 \\
    &{\ruQasper} & - & 7.4 & 0.0 & 0.0 & - & - & 2.5\\
    &{\ruGsm}  & - & - & 0.0 & - & - & - & 0.0 \\
    \bottomrule
    \end{tabular}
  \caption{The table presents the evaluation results of Saiga-LLaMA-3. \textbf{Dataset Name} shows the name of the dataset.
    The rows \textbf{4k}, \textbf{8k}, \textbf{16k}, \textbf{32k}, \textbf{64k}, \textbf{128k} show evaluation scores of datasets for each context length, respectively. The \textbf{Overall} score is obtained by averaging the results over each length.
} \label{tab:result_saiga_llama3}
\end{table*}
\begin{table*}
\small
  \centering
  \renewcommand\arraystretch{0.95}
\begin{tabular}{@{}cp{0.20\linewidth}lccccccc@{}}
    \toprule
    &\textbf{Dataset Name} & \textbf{4k} & \textbf{8k} & \textbf{16k} & \textbf{32k} & \textbf{64k} & \textbf{128k} & \textbf{Overall} \\
    \midrule
    \multirow{2}{*}{\rotatebox[origin=c]{90}{\textbf{I}}}
    &{\passkey} & 100.0 & 100.0 & 0.0 & 0.0 & 0.0 & 0.0 & 33.3 \\
    &{\passkeyWithoutGarbage} & 100.0 & 100.0 & 0.0 & 0.0 & 0.0 & 0.0 & 33.3 \\
    \midrule
    \multirow{8}{*}{\rotatebox[origin=c]{90}{\textbf{II}}}
    &{\matreshkaNames} & 40.0 & 20.0 & 0.0 & 0.0 & 0.0 & 0.0 & 10.0 \\
    &{\matreshkaYesNo} & 62.2 & 59.0 & 0.0 & 0.0 & 0.0 & 0.0 & 20.2 \\
    &{\librusecHistory} & - & 90.6 & 0.0 & 0.0 & 0.0 & - & 22.7 \\
    &{\ruTrec} & 37.8 & 38.0 & 0.0 & 0.0 & - & - & 19.0 \\
    &{\ruSciFi} & - & - & - & 0.0 & 0.0 & - & 0.0 \\
    &{\ruSciAbstractRetrieval} & 97.1 & 88.1 & 0.0 & 0.0 & 0.0 & 0.0 & 30.9 \\
    &{\ruTPO} & - & 58.2 & - & - & - & - & 58.2 \\
    &{\ruQuality} & - & 17.1 & 0.0 & - & - & - & 8.5 \\
    \midrule
    \multirow{8}{*}{\rotatebox[origin=c]{90}{\textbf{III}}}
    &{\LongContextMultiQ} & 9.5 & 32.5 & 0.0 & 0.0 & 0.0 & 0.0 & 7.0 \\
    &{\librusecMHQA} & -  & 41.4 & - & - & - & - & 41.4\\
    &{\wikimqa} & - & 55.1 & 0.0 & 0.0 & - & - & 18.4 \\
    &{\qaFirst} & 68.0 & 57.0 & 0.0 & 0.0 & 0.0 & 0.0 & 20.8 \\
    &{\qaSecond} & 27.0 & 19.0 & 0.0 & 0.0 & 0.0 & 0.0 & 7.7 \\
    &{\qaThird} &  28.5 & 25.9 & 0.0 & 0.0 & 0.0 & 0.0 & 9.1 \\
    &{\qaFourth} & 58.4 & 56.4 & 0.0 & 0.0 & 0.0 & 0.0 & 19.1 \\
    &{\qaFifth} & 67.2 & 68.7 & 0.0 & 0.0 & 0.0 & 0.0 & 22.6 \\
    \midrule
    \multirow{3}{*}{\rotatebox[origin=c]{90}{\textbf{IV}}}
    &{\ruSciPassageCount} & 15.0 & 5.0 & 0.0 & 0.0 & 0.0 & 0.0 & 3.3 \\
    &{\ruQasper} & - & 6.5 & 0.0 & 0.0 & - & - & 2.2\\
    &{\ruGsm}  & - & - & 0.0 & - & - & - & 0.0  \\
    \bottomrule
    \end{tabular}
  \caption{The table presents the evaluation results of LLaMA-3-3B. \textbf{Dataset Name} shows the name of the dataset.
    The rows \textbf{4k}, \textbf{8k}, \textbf{16k}, \textbf{32k}, \textbf{64k}, \textbf{128k} show evaluation scores of datasets for each context length, respectively. The \textbf{Overall} score is obtained by averaging the results over each length.
} \label{tab:result_llama3}
\end{table*}
\begin{table*}
\small
  \centering
  \renewcommand\arraystretch{0.95}
\begin{tabular}{@{}cp{0.20\linewidth}lccccccc@{}}
    \toprule
    &\textbf{Dataset Name} & \textbf{4k} & \textbf{8k} & \textbf{16k} & \textbf{32k} & \textbf{64k} & \textbf{128k} & \textbf{Overall} \\
    \midrule
    \multirow{2}{*}{\rotatebox[origin=c]{90}{\textbf{I}}}
    &{\passkey} & 100.0 & 97.5 & 12.5 & 0.0 & 0.0 & 0.0 & 35.0 \\
    &{\passkeyWithoutGarbage} & 100.0 & 100.0 & 30.0 & 0.0 & 0.0 & 0.0 & 38.3 \\
    \midrule
    \multirow{8}{*}{\rotatebox[origin=c]{90}{\textbf{II}}}
    &{\matreshkaNames} & 32.7 & 16.0 & 0.0 & 0.0 & 0.0 & 0.0 & 8.1 \\
    &{\matreshkaYesNo} & 50.2 & 50.0 & 0.3 & 0.0 & 0.0 & 0.0 & 16.8\\
    &{\librusecHistory} & -  & 78.1 & 15.6 & 0.0 & 0.0 & - & 23.4 \\
    &{\ruTrec} & 2.7 & 10.0 & 3.3 & 0.0 & - & - & 4.0 \\
    &{\ruSciFi} & - & - & - & 2.8 & 0.0 & - & 1.4\\
    &{\ruSciAbstractRetrieval} & 94.8 & 76.1 & 0.0 & 0.0 & 0.0 & 0.0 & 28.5 \\
    &{\ruTPO} & - & 39.6 & - & - & - & - & 39.6\\
    &{\ruQuality} & - & 22.8 & 11.8 & - & - & - & 17.3 \\
    \midrule
    \multirow{8}{*}{\rotatebox[origin=c]{90}{\textbf{III}}}
    &{\LongContextMultiQ} & 4.0 & 22.0 & 0.5 & 0.0 & 0.0 & 0.0 & 4.4 \\
    &{\librusecMHQA} & - & 34.1 & - & - & - & - & 34.1\\
    &{\wikimqa} & - & 42.9 & 18.0 & 8.1 & - & - & 23.0 \\
    &{\qaFirst} & 63.0 & 63.0 & 0.0 & 0.0 & 0.0 & 0.0 & 21.0 \\
    &{\qaSecond} & 21.0 & 25.0 & 0.0 & 0.0 & 0.0 & 0.0 & 7.7  \\
    &{\qaThird} &  29.0 & 25.0 & 0.0 & 0.0 & 0.0 & 0.0 & 9.0 \\
    &{\qaFourth} & 42.9 & 31.6 & 0.0 & 0.0 & 0.0 & 0.0 & 12.4 \\
    &{\qaFifth} & 70.0 & 69.3 & 0.0 & 0.0 & 0.0 & 0.0 & 23.2 \\
    \midrule
    \multirow{3}{*}{\rotatebox[origin=c]{90}{\textbf{IV}}}
    &{\ruSciPassageCount} & 4.0 & 4.0 & 0.0 & 0.0 & 0.0 & 0.0 & 1.3 \\
    &{\ruQasper} & - & 6.3 & 1.1 & 0.1 & - & - & 2.5\\
    &{\ruGsm}  & - & - & 13.0 & - & - & - & 13.0 \\
    \bottomrule
    \end{tabular}
  \caption{The table presents the evaluation results of Mistral-7B-V0.1. \textbf{Dataset Name} shows the name of the dataset.
    The rows \textbf{4k}, \textbf{8k}, \textbf{16k}, \textbf{32k}, \textbf{64k}, \textbf{128k} show evaluation scores of datasets for each context length, respectively. The \textbf{Overall} score is obtained by averaging the results over each length.
} \label{tab:result_mistral_7b_v01}
\end{table*}

\end{appendices}
\end{document}